\documentclass{article}

\usepackage[accepted]{icml2026}

\usepackage{mathtools}    
\usepackage{amssymb}
\usepackage{bm}

\usepackage{times}
\usepackage{textcomp}
\usepackage{soul}
\usepackage{graphicx}
\usepackage{wrapfig}
\usepackage{float}
\usepackage{epstopdf}     
\usepackage{adjustbox}

\usepackage{booktabs}
\usepackage{tabularx}
\usepackage{multirow}
\usepackage{makecell}
\usepackage{colortbl} 
\usepackage{hhline}
\usepackage{siunitx,ragged2e}
\usepackage{tcolorbox}
\usepackage[table,xcdraw,dvipsnames]{xcolor}
\usepackage{dashrule}
\usepackage{caption}
\usepackage{subcaption}

\usepackage{natbib}
\setcitestyle{authoryear,open={(},close={)}}

\usepackage{enumitem}
\usepackage{enumerate}
\usepackage{pifont}
\usepackage{bbding}
\usepackage{comment}
\usepackage{fancyvrb}
\usepackage{blindtext}
\usepackage{setspace}
\usepackage{url}
\usepackage[english]{babel}
\usepackage{wrapfig}

\definecolor{citepink}{RGB}{185,80,135}
\usepackage[colorlinks=true,
            citecolor=citepink]{hyperref}
\hypersetup{colorlinks}
\usepackage{cleveref}

\usepackage{newtxtext,newtxmath}

\usepackage{pgfplots}
\pgfplotsset{compat=1.18}

\usepackage{enumitem}
\usepackage{tikz}

\definecolor{llmfree}{RGB}{180,180,180}
\definecolor{llmdriven}{RGB}{120,170,255}
\definecolor{retrieval}{RGB}{70,130,200}

\definecolor{StableBlueStrong}{RGB}{30, 110, 180}
\definecolor{softblue}{RGB}{60,90,180}
\definecolor{darkpastelgreen}{rgb}{0.01, 0.75, 0.24}
\definecolor{electriccrimson}{rgb}{1.0, 0.0, 0.25}
\definecolor{navyblue}{rgb}{0.0, 0.0, 0.75}
\definecolor{gptfivegreen}{rgb}{0.0,0.45,0.0} 

\newcommand{\algacro}{{CUA-Skill}}
\newcommand{\cuaskillagent}{{CUA-Skill Agent}}

\icmltitlerunning{\algacro{}: Developing Computer Using Agents with a Skill Framework}

\begin{document}

\twocolumn[
  \icmltitle{\algacro{}: Develop Skills for Computer Using Agent}



  \icmlsetsymbol{equal}{*}

  \begin{icmlauthorlist}
  	\icmlauthor{Tianyi Chen}{equal,msft}
   	\icmlauthor{Yinheng Li}{equal,msft}
    \icmlauthor{Michael Solodko}{equal,msft}
    \icmlauthor{Sen Wang}{equal,msft}
    \icmlauthor{Nan Jiang}{msft}
   	\icmlauthor{Tingyuan Cui}{msft}
   	\icmlauthor{Junheng Hao}{msft}
   	\icmlauthor{Jongwoo Ko}{msft}
   	\icmlauthor{Sara Abdali}{msft}
   	\icmlauthor{Qing Xiao}{msft}
   	\icmlauthor{Leon Xu}{msft}
   	\icmlauthor{Suzhen Zheng}{msft}
   	\icmlauthor{Hao Fan}{msft}
   	\icmlauthor{Pashmina Cameron}{msft}
   	\icmlauthor{Justin Wagle}{msft}
   	\icmlauthor{Kazuhito Koishida}{msft}
  \end{icmlauthorlist}
  \icmlaffiliation{msft}{Microsoft, author roles in Appendix~\ref{appendix:authorlist}}



  \vskip 0.3in
]

\printAffiliationsAndNotice{\icmlEqualContribution}


\begin{abstract}
Computer-Using Agents (CUAs) aim to autonomously operate computer systems to complete real-world tasks. However, existing agentic systems remain difficult to scale and lag behind human performance. A key limitation is the absence of reusable and structured skill abstractions that capture how humans interact with graphical user interfaces and how to leverage these skills. We introduce \algacro{}, a computer-using agentic skill base that encodes human computer-use knowledge as skills coupled with parameterized execution and composition graphs.
\algacro{} is a large-scale library of carefully engineered skills spanning common Windows applications, serving as a practical infrastructure and tool substrate for scalable, reliable agent development.
Built upon this skill base, we construct \cuaskillagent{}, an end-to-end computer-using agent that supports dynamic skill retrieval, argument instantiation, and memory-aware failure recovery. Empirically, \algacro{} substantially improves the quality and reliability of trajectory generation, achieving a 76.4\%  success rate, which is multiple times higher than existing baselines. On the challenging end-to-end WindowsAgentArena benchmark, \cuaskillagent{} further attains state-of-the-art performance with a 57.5\% best-of-three success rate, while remaining significantly more efficient than prior and concurrent approaches. Together, \algacro{} serves as a strong and scalable foundation for building future CUA systems. The project page is available at \url{https://microsoft.github.io/cua_skill/}.
\end{abstract}

\section{Introduction}

\begin{figure}[t]
	\centering
	\includegraphics[width=\linewidth]{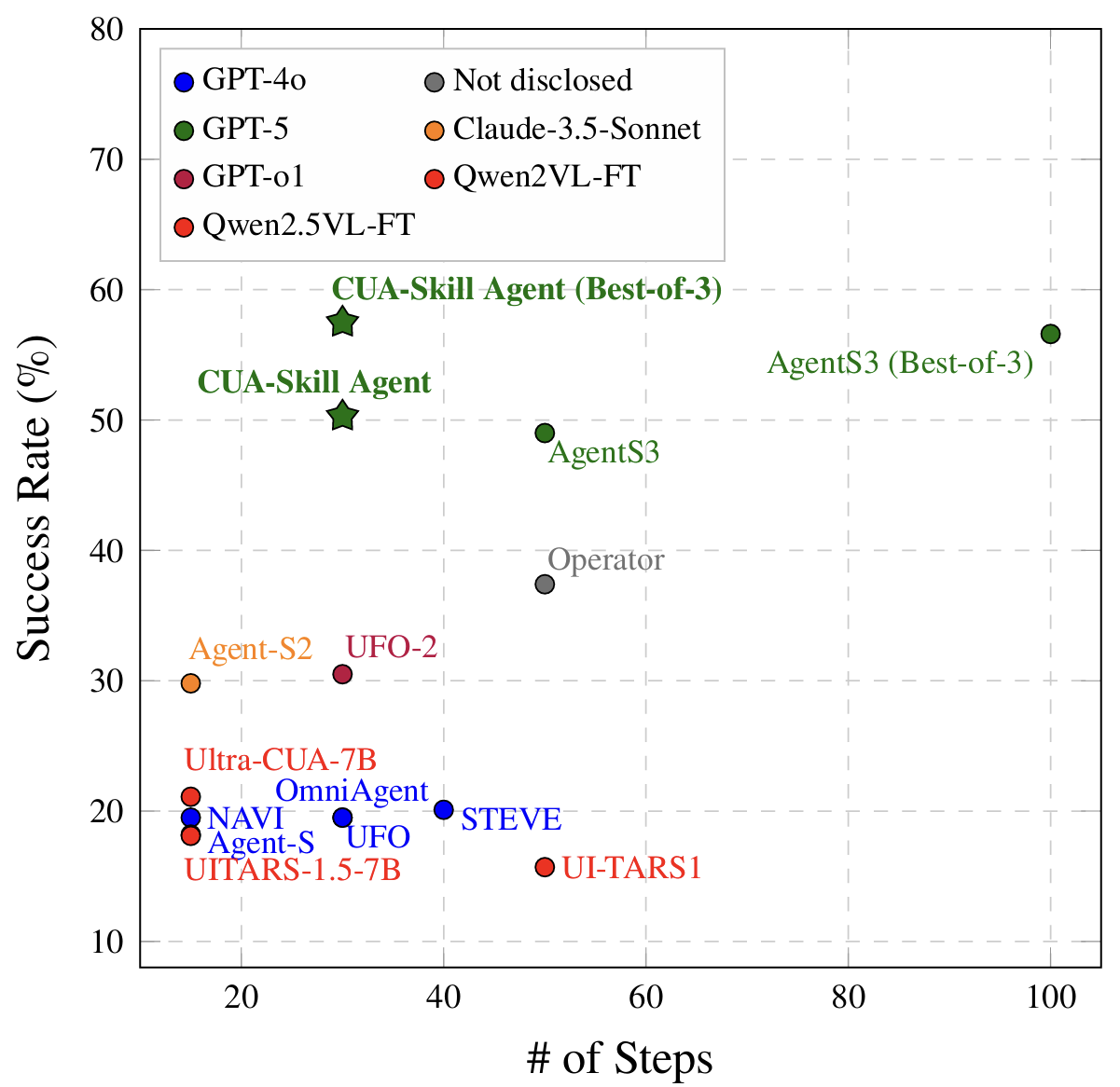}
	\captionsetup{skip=2pt} 
	\caption{Success rate vs.\ execution steps on WAA.}
	\label{fig:waa_steps_vs_success}
	\vspace{-1.5em}
\end{figure}

Computer-Using Agents (CUAs) aim to autonomously operate graphical user interfaces (GUIs) to complete real-world desktop tasks such as document editing, web navigation, data analysis, and system configuration~\citep{xie2024osworld,zhang2025phi,yang2025ultracua,hui-etal-2025-winspot}. Recent advances in large language models (LLMs) and multimodal perception have substantially improved agents’ abilities to interpret user intent and visually ground actions on the screen, making CUAs a promising pathway toward general-purpose digital assistants capable of interacting with complex desktop environments.

\begin{figure*}[ht]
	\centering
	\fbox{\includegraphics[width=0.95\textwidth]{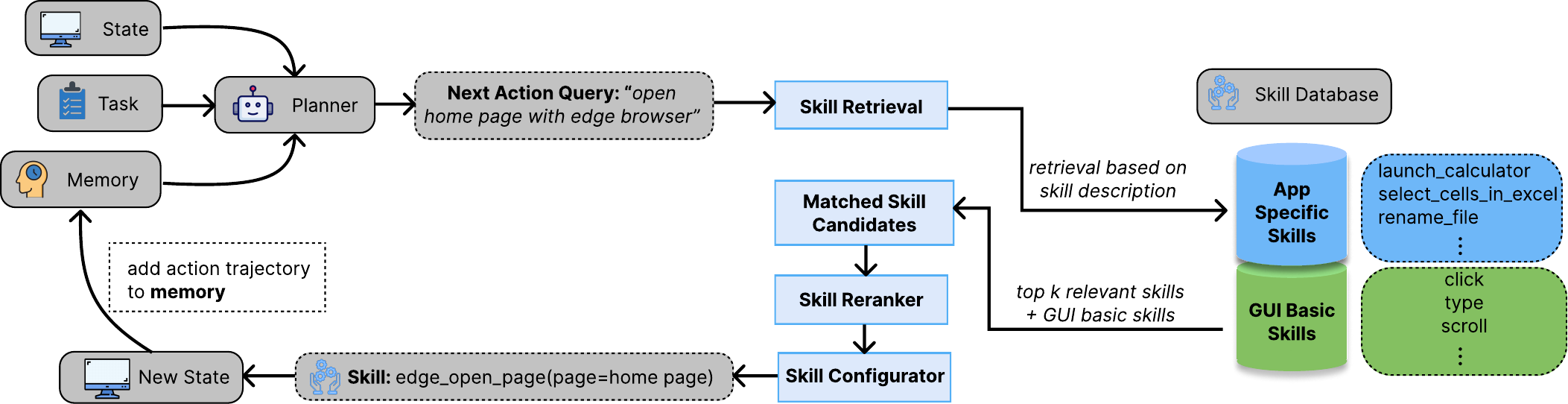}}
	\caption{Overview of CUA-Skill and Associated Skill-Agent.}
	\label{fig:overview}
	\vspace{-1em}
\end{figure*}

Despite this progress, building reliable and scalable CUAs remains challenging. Existing systems often struggle with long-horizon tasks that require executing dozens of interdependent actions across dynamic UI states. Small errors in grounding, planning, or execution can quickly compound, leading to brittle behavior and low end-to-end success rates. More fundamentally, most current approaches lack an explicit representation of \emph{how humans use computers}: desktop interaction is typically modeled as flat sequences of low-level actions, forcing agents to repeatedly rediscover common workflows from scratch.

In contrast to current CUAs, human computer use is inherently structured around reusable procedural knowledge. Users rely on familiar \emph{skills}, such as launching applications, navigating menus, or formatting documents, which are composed into higher-level workflows and adapted to the current UI context. The lack of such reusable and structured skill abstractions remains a key bottleneck for existing CUAs, limiting their scalability, generalization, and robustness on complex real-world tasks.

Concurrently, Anthropic introduced the notion of “agent skills” as reusable, filesystem-based resources that encapsulate domain expertise \citep{claudeskill}. While effective in code-centric environments (e.g., Linux or API-rich systems), these skills are primarily executed through scripts and tightly integrated with the Model Context Protocol (MCP) \citep{anthropic2024mcp}. As a result, they are less suited for desktop environments such as Windows, where many applications expose limited or inconsistent programmatic APIs and effective task execution fundamentally makes them difficult to leverage across applications. Therefore, a question is naturally raised:

\emph{\textbf{\textcolor{softblue}{How can we build a scalable and transferable {skill base} for desktop environments that captures human procedural knowledge and enables reliable and capable CUAs?}}}

In this work, we answer this question by introducing \algacro{}, the first systematic agentic skill library designed for desktop computer use. \algacro{} encodes human computer-use knowledge as reusable \emph{skills} coupled with \textbf{parameterized execution and composition graphs}, forming a structured intermediate layer between high-level user intent and low-level interaction primitives. 
While GUI primitives serve as the primary, human-aligned substrate for skill execution, the execution graph abstraction flexibly supports script- and code-based execution paths when they offer improved reliability or efficiency. 
This unified parameterization makes skills transferable across tasks, UI states, and applications, enabling strong generalization.

Built on top of this skill base, we develop \textbf{\cuaskillagent{}}, an end-to-end computer-using agent that performs retrieval-augmented skill selection, configuration, and execution. At each step, an LLM-based planner retrieves relevant skills conditioned on the current UI state and user goal, re-ranks candidates using execution context and memory, instantiates skill arguments, and executes the selected skill via GUI grounding or direct script execution, depending on the instantiated execution path. This design supports scalable skill expansion, memory-aware recovery from failures, and robust long-horizon task completion without hard-coding tools into prompts or relying on monolithic plans~\citep{huang2023metatool,schick2023toolformer}.

Our main contributions are summarized as follows:
\begin{itemize}[leftmargin=1em]
	\item \textbf{\algacro{}.} We introduce a structured agentic skill library for desktop environments that encodes human computer-use knowledge as reusable, parameterized skills with explicit execution and composition graphs. This design enables strong transferability and generalization across tasks and UI states. The initial release contains hundreds of carefully engineered atomic skills spanning tens of popular applications. Through parameterization and composition, these skills can be instantiated into millions or more executable task variants, supporting a wide range of downstream agent applications.
	
	\item \textbf{\cuaskillagent{}.} To effectively utilize \algacro{}, we propose a skill-centric, retrieval-augmented agent that performs dynamic skill retrieval, argument instantiation, and execution. The agent supports scalable skill expansion, memory-aware failure recovery, and robust long-horizon desktop task completion.
	
	\item \textbf{State-of-the-Art Performance.} Extensive evaluations demonstrate that \algacro{} substantially improves the performance of multiple agent applications. In the trajectory generation, \algacro{} achieves a \textbf{76.4\%} success rate, outperforming existing approaches by $\bm{1.7\times-3.6\times}$. On the end-to-end WindowsAgentArena CUA benchmark, \cuaskillagent{} attains state-of-the-art results, achieving a best-of-three success rate of \textbf{57.5\%}.
\end{itemize}

\section{Related Works}\label{sec:related_work}

In this section, we discuss two related topics to \algacro{}. More related works are provided in Appendix~\ref{appendix:more_related_works}.

\paragraph{Memory Modules and Knowledge Graph Integration.}
Memory modules are a core architectural component in many recent computer-use agents (CUAs), enabling agents to track task progress and reuse historical information across long-horizon interactions~\citep{agashe2025agents2compositionalgeneralistspecialist, song2025coact1computerusingagentscoding, wang2025uitars2technicalreportadvancing}. By retaining execution history and intermediate outcomes, memory supports more informed planning decisions and mitigates the limitations of a single prompt window~\citep{park2023generativeagents}. Recent systems further structure memory as explicit graphs with retrieval and update operations, allowing agents to store intermediate facts, revise beliefs, and incorporate new evidence over time~\citep{chhikara2025mem0}. In parallel, recent frameworks emphasize iterative query, update interactions with knowledge graphs to improve reasoning consistency and reduce hallucination, with demonstrated benefits in mobile-agent settings~\citep{guan2025kgrag}.

\textcolor{navyblue}{\textbf{\textit{Relation to \algacro{}.}}}
These approaches primarily focus on modeling \emph{what the agent knows}, i.e., task state, observations, and historical outcomes, but do not explicitly encode \emph{how humans perform computer interactions} as reusable procedures. In particular, they lack action-level abstractions with parameterized execution semantics, limiting the systematic reuse of interaction knowledge across tasks, applications, and UI contexts. \textcolor{softblue}{In contrast, \algacro{} targets this missing procedural layer by encoding human computer-use behavior as reusable skills with \textbf{parameterized execution graphs, enabling transferable and reliable procedural knowledge for desktop environments}}.

\paragraph{Structured Task Planning and MCPs.}
Recent work on tool-using agents increasingly frames computer use as a structured planning problem, where success depends on coordinating actions over long horizons rather than selecting isolated tool calls~\citep{zhuang2023toolqa, chen2025appselectbench}. Code-based planning further strengthens such compositions by compiling high-level intents into modular units that can be reused across tasks~\citep{singh2022progprompt}. Desktop-agent foundations emphasize that explicit workflow structures become increasingly important as task horizons grow, as they support state tracking and recovery under tool and UI ambiguity~\citep{wang2025opencua}.

In parallel, recent standardization efforts such as the Model Context Protocol (MCP) focus on unifying \emph{how} agents access external software, tools, and data sources through standardized client--server interfaces~\citep{anthropic_mcp_2024,mcp_spec_2025}. While MCP and related tool abstractions provide a powerful foundation for connecting agents to software systems, they typically require substantial engineering effort to expose application-specific APIs and maintain underlying codebases, especially for complex desktop environments.

\textcolor{navyblue}{\textbf{\textit{Relation to \algacro{}.}}}
\algacro{} extends both structured planning frameworks and MCP-style tool interfaces, but introduces a new abstraction level to fulfill the gaps. Instead of requiring deep software integration or tool implementations, \algacro{} encodes human computer-use knowledge as reusable skills with parameterized execution and composition graphs. 
\algacro{} significantly lowers the engineering burden for skill authors and agent developers. \textcolor{softblue}{\textbf{This design makes \algacro{} user-friendly to construct, easier to extend and maintain, and naturally reusable across applications and tasks.}} As a result, \algacro{} provides a practical and scalable substrate for robust desktop CUA.

\section{Computer-Using Agentic Skills}\label{sec:cua_skill}

This section introduces \textbf{\algacro{}}, a structured and parameterized skill abstraction system designed to encode {human computer-use knowledge} for desktop environments. Our core premise is that effective computer use is not a flat sequence of primitive GUI actions, but a composition of reusable, intent-aligned skills that humans routinely apply across tasks and applications, each admitting multiple valid realizations under different states.

Formally, \algacro{} consists of three components: (\emph{i}) a \emph{skill cell} that captures minimal user intent, (\emph{ii}) a \emph{parameterized execution graph} that specifies concrete realizations of the skill through the ways like GUI-grounded interactions and executable scripts, and (\emph{iii}) a \emph{composition graph} that encodes how individual skills are typically chained together.

\subsection{Skill}\label{sec:atomic_skill}

Skill is the primitive behavioral units in \algacro{}. Each skill is denoted by ${S}$ and captures a minimal but meaningful user intent. The collection of skills is denoted as $\mathcal{S}$. 
\begin{equation}\label{eq:skill_def}
	S:=\{\tau, \mathcal{I}, \mathcal{A}, \mathcal{G}_e\}.    
\end{equation}
A skill $S$ is defined by, \textit{(i)} a suitable application $\tau$, \textit{(ii)} a natural language user intent $\mathcal{I}$, \textit{(iii)} an argument pool $\mathcal{A}$, and \textit{(iv)} a parameterized execution graph $\mathcal{G}_e$. 
The argument schema $\mathcal{A}=\{A_1,\cdots,A_K\}$ specifies a set of type slots that describe the information that the skill needs from the user or the environment. The execution graph $\mathcal{G}_e$ encodes how to realize the intent as a sequence of low-level interactions, such as keystrokes, mouse events, or application-specific API calls, conditional on those arguments. By constraining skills to be small and application-specific, \algacro{} can reliably reuse them as building blocks when constructing longer multi-step workflows across applications.

\paragraph{Feasible Domain and Generator for Argument}
For each argument $A \in \mathcal{A}$, we associate a \emph{feasible domain} $\mathcal{D}(A)$ that specifies the set of values for which the skill remains well-defined and executable. These domains are  defined as part of the skill specification and reflect both application semantics and desktop environment constraints.

We distinguish between two broad categories of argument domains. \textcolor{ForestGreen}{\textbf{Finite-Domain Arguments}} correspond to discrete and enumerable choices, such as menu items, toolbar options, system toggles, or predefined configuration states. For such arguments, $\mathcal{D}(A)$ is a finite set that can be exhaustively enumerated or dynamically queried from the UI state. In contrast, \textcolor{ForestGreen}{\textbf{Open-Domain Arguments}} correspond to unbounded or high-cardinality inputs, such as file paths, textual content, and numerical values, etc. These domains are typically infinite or impractically large to enumerate and require structured sampling strategies.

The feasible domain definition enables CUA-Skill to reason about argument validity independently of execution, ensuring that each instantiated skill corresponds to a realizable interaction on the desktop. Moreover, feasible domains allow us to associate each argument type with a specialized argument generator, tailored to the structure of $\mathcal{D}(A)$. For example, finite-domain arguments may be instantiated via enumeration or UI-state grounding, while open-domain arguments may be generated through controlled sampling, or environment-aware heuristics.

\begin{figure}[h]
	\centering
	\includegraphics[height=0.25\textheight]{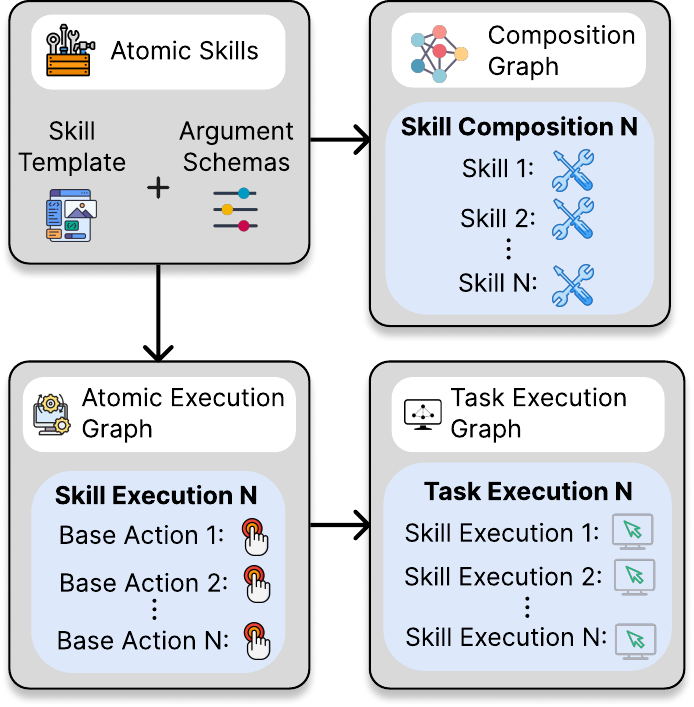}
	\caption{
		\textbf{CUA Skill and Graph Construction.}
	}
	\label{fig:excel_execution_graph}
\end{figure}

\begin{algorithm}[tb]
	\caption{\cuaskillagent{}}\label{alg:cua_skill_agent}
	\begin{algorithmic}[1]
		\STATE {\bfseries Input:} User instruction $\mathcal{U}$, planner $\mathcal{M}_p$, retrieval module $\mathcal{R}$ over skill collection $\mathcal{S}$, basic skill subset $\mathcal{S}_\text{basic} \subseteq \mathcal{S}$, memory $\mathcal{M}$, environment $\mathbb{E}$.
		\STATE {\bfseries Hyperparameters:} query budget $K$, skill budget $L$.
		\STATE Initialize memory $\mathcal{M} \gets \{\mathcal{U}\}$ and timestamp $t \gets 0$.
		\WHILE{termination condition is not satisfied}
		\STATE Obtain observation $o_t \gets \text{GetObservation}(\mathbb{E}).$
		\STATE \textbf{Query generation:} LLM produces $K$ queries
		$$
		Q_t \gets \text{QueryGenerator}(\mathcal{M}_p, \mathcal{U}, o_t, \mathcal{M}, K).
		$$
		\vspace{-1em}
		\STATE \textbf{Skill retrieval:} retrieve top-$L$ candidates from $\mathcal{S}$.
		$$
		\mathcal{C}_t \gets \text{RetrieveTopLQuery}(\mathcal{R}, Q_t, S, T).
		$$
		\vspace{-1em}
		\STATE \textbf{Skill re-ranking:} pick the most promising skill, considering both retrieved and basic skills
		$$
		S_t \gets \text{SkillReranker}(\mathcal{M}_p, \mathcal{C}_t \cup \mathcal{S}_\text{basic}, o_t, \mathcal{M}).
		$$
		\vspace{-1em}
		\STATE \textbf{Skill configuration:} configure arguments
		$$
		\hat{S}_t \gets \text{SkillConfigurator}(\mathcal{M}_p, S_t, o_t, \mathcal{M}).\label{line:skill_config}
		$$
		\vspace{-1em}
		\STATE \textbf{Skill execution} (call grounder model if needed)
		$$
		\text{outcome}_t \gets \text{ExecuteSkill}(\hat{S}_t, \mathbb{E}).
		$$
		\vspace{-1em}
		\STATE \textbf{Update memory:} append skill and outcomes
		$$
		\mathcal{M} \gets \mathcal{M} \cup \{\text{Summarize}(\hat{S}_t, \text{outcome}_t)\}.
		$$
		\vspace{-1em}
		\STATE $t \gets t + 1$.
		\ENDWHILE
	\end{algorithmic}
\end{algorithm}

\subsection{Skill Execution Graph}

For each skill ${S} \in \mathcal{S}$, we construct a \emph{skill execution graph} $\mathcal{G}_e(S) = (\mathcal{V}, \mathcal{E})$ that provides one or more concrete procedures for realizing the user intent. Each execution graph may comprise {GUI-grounded interaction primitives} or {executable script actions}, unified within a single representation. Unlike a fixed action sequence, the execution graph encodes a parameterized structured space of valid interaction paths that account for common UI variations, alternative execution realizations, and execution contingencies.

Each node $v \in \mathcal{V}$ represents an internal control state of the skill, including a designated start state and one or more terminal states. Each directed edge $(v, a, v') \in \mathcal{E}$ is labeled by a \emph{base action} $a$, which may correspond to a GUI interaction primitive or an executable script action, and may be guarded by UI predicates that condition execution on the current screen state. The execution graph is parameterized by a concrete argument instantiation from $\mathcal{D}(A_1) \times \cdots \times \mathcal{D}(A_K)$, which determines concrete interaction targets, such as UI elements, file paths, or input content. Concrete examples are present in Appendix~\ref{appendix:cua_execution_graph}.

In practice, most execution graphs are compact directed graphs with a dominant execution path and a small number of guarded branches. These branches handle common UI variants, such as alternative menu layouts, dialog prompts, or multiple valid interaction affordances, enabling skills to remain robust to UI changes without requiring redefinition. Moreover, the execution graph supports edge weighting mechanisms that encode execution preferences for different downstream use cases or human preferences.

\subsection{Skill Composition Graph}

The {skill composition graph} is a directed graph $\mathcal{G}_c = (\mathcal{V}_c, \mathcal{E}_c)$ that encodes how individual skills can be composed into higher-level user tasks. Each node $v \in \mathcal{V}_c$ corresponds to a skill ${S}_v$, and each directed edge $(u, v) \in \mathcal{E}_c$ represents a valid composition from skill ${S}_u$ to skill ${S}_v$.

A path $(v_1, \ldots, v_T)$ in $\mathcal{G}_c$ defines a multi-step task workflow, where nodes represent intermediate sub-goals and edges encode ordering and compatibility constraints between skills. Importantly, the skill composition graph captures reusable procedural knowledge about how skills are typically chained in human computer use, rather than prescribing a fixed execution plan.

We organize $\mathcal{G}_c$ into single-application and cross-application scenarios. Edge transitions may connect skills within the same application or across different applications. This unified representation allows \algacro{} to model both single-application and multi-application workflows within a shared compositional structure.

\begin{figure}[h]
	\centering
	\includegraphics[height=0.35\textheight]{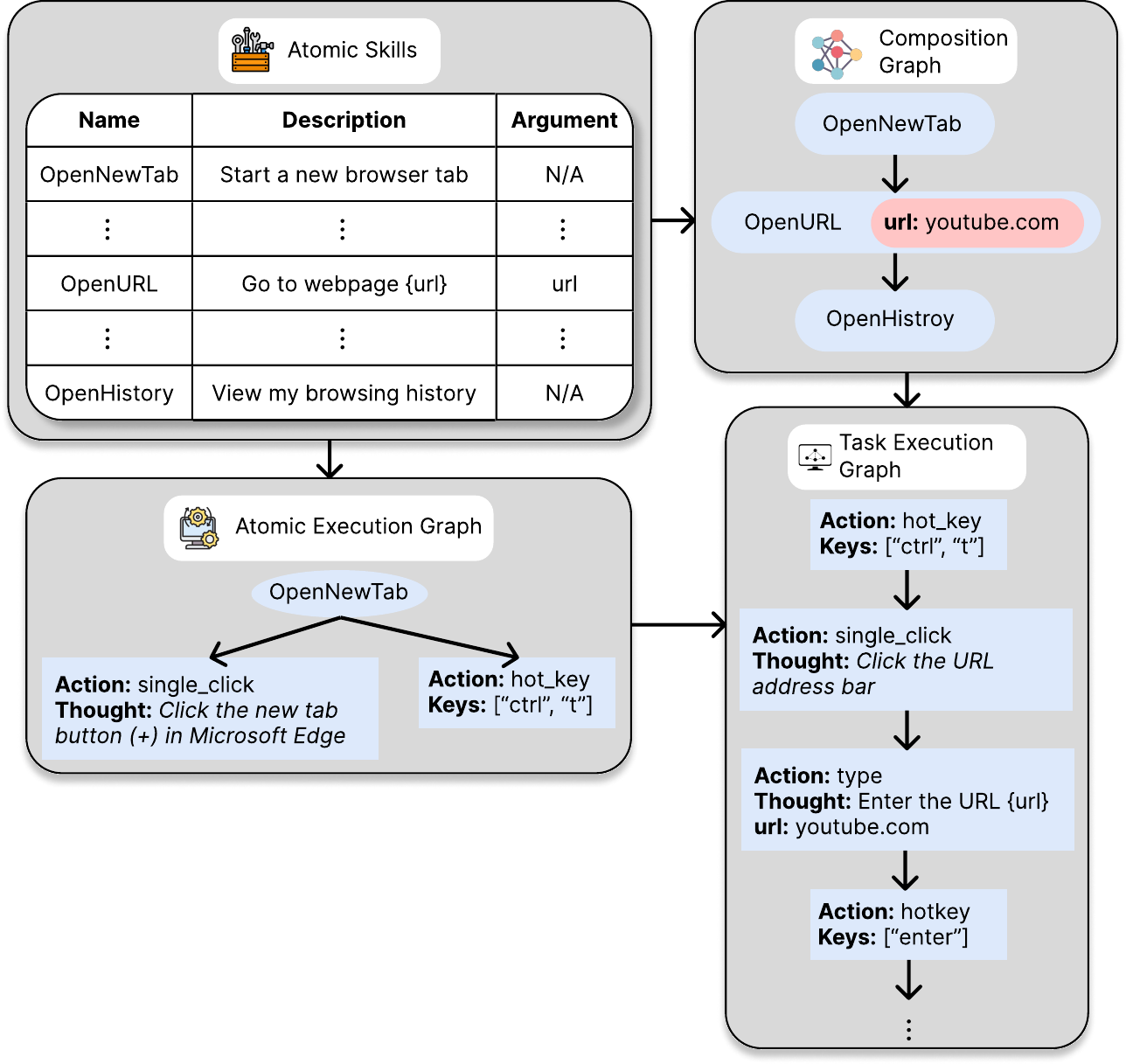}
	\caption{
		\textbf{CUA Skill and Graph Construction Example.}
	}
	\label{fig:execution_graph}
	\vspace{-1em}
\end{figure}

\section{\cuaskillagent{}}\label{sec:cua_skill_agent}

We now design a \textbf{\cuaskillagent{}} that supports flexible, long-horizon task completion via dynamic skill selection and execution. Given a natural-language user instruction, the agent incrementally selects, configures, and executes skills from the \algacro{} library, conditioning each decision on the current UI state, execution history, and accumulated memory. At each step, an LLM planner $\mathcal{M}_p$ determines both \emph{which} skill to invoke next and \emph{how} to instantiate its arguments. This design enables adaptive task completion under UI variability and execution uncertainty.

The overall architecture of \cuaskillagent{} is depicted in \autoref{fig:overview} and stated in Algorithm~\ref{alg:cua_skill_agent}. It consists of five core components: \textit{(i)} a retrieve-augmented skill planner, \textit{(ii)} a skill re-ranker module, \textit{(iii)} a skill argument configuration module, \textit{(iv)} a memory module for to store past action trajectory and execution feedback, and \textit{(v)} an executor.

\subsection{Retrieve-Augmented Skill Planner}

The planner of \cuaskillagent{} is a \emph{Retrieve-Augmented Skill Planner}. Similar while \emph{different} to the tool invoker in MCP~\citep{anthropic2024mcp}, it uses an LLM (e.g., GPT-5) to select an appropriate skill conditioned on the current screen state, execution history, and user goal. Rather than exposing the full skill library to the model context, the planner operates over a structured retrieve-augmented pipeline that narrows the skill space before decision making. The planner $\mathcal{M}_p$ participates throughout the planning process, including skill selection and argument configuration, enabling coherent reasoning across all planning stages.

\subsection{Skill Selection}

\paragraph{Query Generator.}
The Query Generator leverages the planner $\mathcal{M}_p$ to produce candidate retrieval queries for skills that can advance the user goal.
A central challenge is that the LLM has no prior knowledge of the available skill inventory. Although fine-tuning the model with the full skill list is possible, doing so would require retraining whenever new skills are introduced.
Instead, we rely entirely on test-time LLM capabilities and assume that skill names and descriptions follow common natural-language conventions.
Under this assumption, the LLM can generate sufficiently general queries that  match relevant skills via retrieval.

For better selection, we employ two mechanisms: ensembled query generation, where multiple queries with varying wording granularity are generated to cover semantic interpretations (Appendix~\ref{appendix:ensemble_query_generation}); skill reranker that re-evaluates skill candidates and selects the most  appropriate skill.

\begin{table*}[t]
	\centering
	\caption{(Left) Statistics of \algacro{} Execution Graph across applications. The GUI primitive statistics measures per atomic skill, how the quantity of GUI primitives distributes. (Right) Bar plot of success rate across applications.}
	\label{tab:cua_skill_graph_statistic}
	
	\begin{minipage}[t]{0.62\textwidth}
		\vspace{0pt}
		\centering
		\scriptsize
		\begin{tabular}{lccccc}
			\toprule
			\multirow{2}{*}{\textbf{Category}} &
			\multirow{2}{*}{\textbf{\# Atomic Skills}} &
			\multicolumn{2}{c}{\textbf{Action Primitive Statistics}} &
			\multirow{2}{*}{\textbf{Success Rate (\%)}} 
			\\
			\cmidrule(lr){3-4}
			&  &  
			\textbf{\# Mean $\pm$ Std} &
			\textbf{Range} &
			\\
			\midrule
			\rowcolor{blue!8}
			\multicolumn{5}{c}{\textbf{Basic \& Common GUI Primitives}} \\
			\midrule
			Basic Primitives & 29 & $1.00\ \pm\ 0.00$ & [1--1] & 94\% \\
			\midrule
			
			\rowcolor{blue!8}
			\multicolumn{5}{c}{\textbf{Application-Level Atomic Skill Distribution}} \\
			\midrule
			Amazon            & 20 & $2.40\ \pm\ 2.22$ & [1--9]  & 50\% \\
			Bing Search       & 19 & $3.20\ \pm\ 1.10$ & [1--4]  & 94\% \\
			Calculator        & 33 & $1.90\ \pm\ 0.69$ & [1--3]  & 82\% \\
			Clock             & 20 & $3.70\ \pm\ 3.38$ & [1--20] & 75\% \\
			Excel             & 18 & $4.40\ \pm\ 5.38$ & [1--9]  & 100\% \\
			File Explorer     & 50 & $2.10\ \pm\ 2.40$ & [1--6]  & 72\% \\
			Google Chrome     & 31 & $4.10\ \pm\ 1.17$ & [1--12] & 82\% \\
			Microsoft Edge    & 38 & $5.20\ \pm\ 3.19$ & [1--16] & 80\% \\
			Notepad           & 33 & $5.10\ \pm\ 4.38$ & [1--20] & 70\% \\
			Paint             & 7  & $6.70\ \pm\ 1.80$ & [3--9]  & 78\% \\
			PowerPoint        & 45 & $3.80\ \pm\ 2.02$ & [1--9]  & 70\% \\
			VLC Player        & 26 & $3.70\ \pm\ 3.02$ & [1--13] & 60\% \\
			VS Code           & 20 & $2.90\ \pm\ 1.29$ & [1--7]  & 73\% \\
			Windows Settings  & 21 & $2.00\ \pm\ 0.97$ & [1--4]  & 75\% \\
			Word              & 42 & $3.60\ \pm\ 2.09$ & [1--9]  & 70\% \\
			YouTube           & 26 & $3.70\ \pm\ 1.21$ & [1--7]  & 72\% \\
			\midrule
			
			\textbf{Total} &
			\textbf{478}  &
			\textbf{$3.75\ \pm\ 2.91$} &
			[1-20] &
			\textbf{76.4\%}\\
			\bottomrule[0.1em]
		\end{tabular}
	\end{minipage}
	\hfill
	\begin{minipage}[t]{0.32\textwidth}
		\vspace{0pt}
		\centering
		\includegraphics[width=\linewidth]{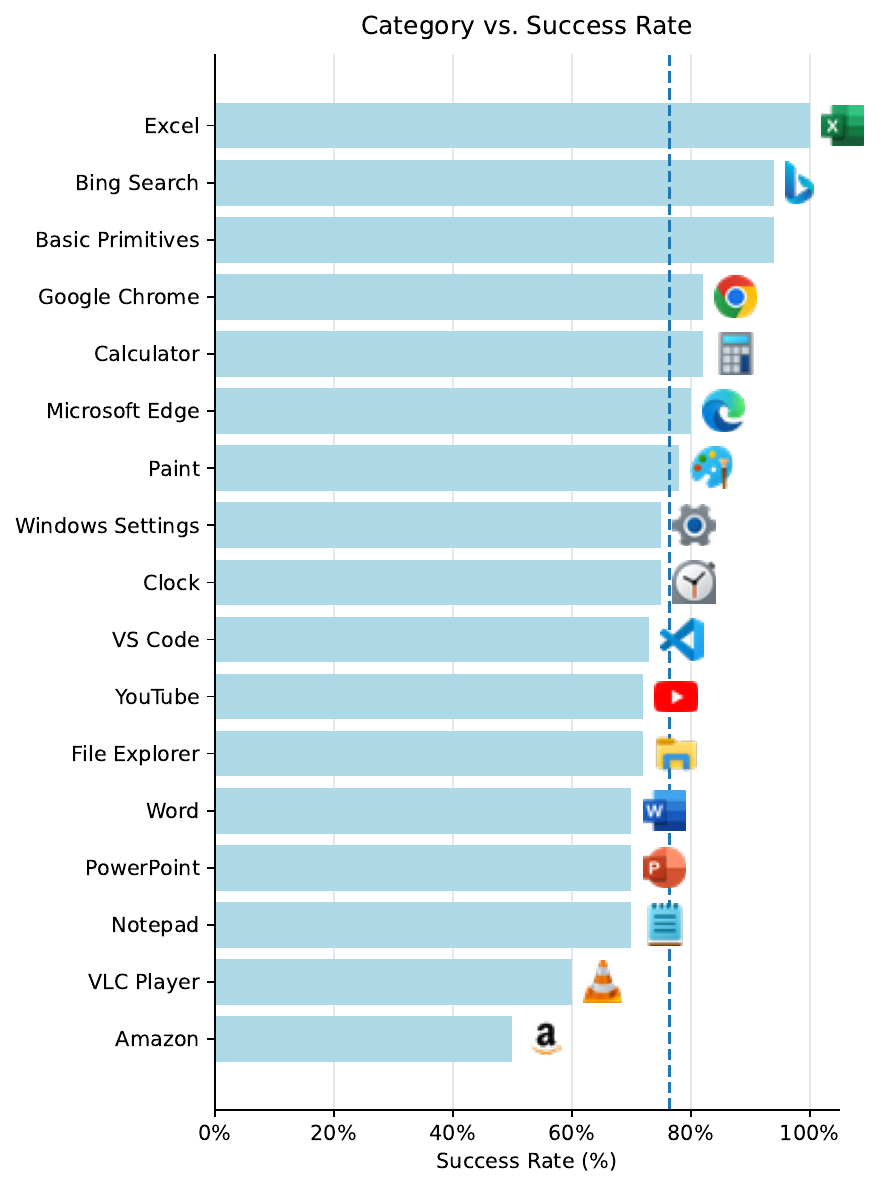}
	\end{minipage}
	\vspace{-1em}
\end{table*}

\paragraph{Skill Retrieval.}

We adopt a hybrid retrieval strategy that combines lexical matching with semantic retrieval, as such hybrids have shown strong performance in text retrieval tasks \citep{thakur2021beirheterogenousbenchmarkzeroshot}. In contrast to many MCP architectures that expose the entire tool set directly to model context, \cuaskillagent{} retrieves only a small set of relevant skills, improving scalability and inference efficiency.

During indexing, each skill $S$ is embedded using its name and functional description $\mathcal{I}$. We use Qwen3-Embedding-0.6B~\cite{yang2025qwen3} to construct the semantic index, while maintaining an inverted index over skill text for lexical retrieval. Our architecture is also compatible with other semantic retrieval models. For each generated query, we retrieve by default top-$5$ most relevant skills from both channels and merge them into a consolidated candidate set, which is then passed to the subsequent skill reranker. 

\paragraph{Skill Re-ranker.} The reranker evaluates the retrieved candidate skills and selects the one to make meaningful progress toward the user goal. This evaluation relies on the current UI state, execution history, and the compatibility between candidate skills and their required arguments.

\textbf{Skill Fallback.} Note that in addition to retrieved skills, \textcolor{ForestGreen}{\textbf{the reranker also considers a small set of basic low-level actions}} (e.g., mouse clicks and keyboard actions). This allows the \cuaskillagent{} to fall back to fine-grained control when high-level skills are insufficient, enhancing robustness beyond the predefined skill library.

\subsection{Skill Configurator}

Once a skill $S$ is selected, the planner $\mathcal{M}_p$ instantiates its arguments $\mathcal{A}$ by conditioning on the current UI state, execution history, and user goal. Each argument is generated within its feasible domain, see Section~\ref{sec:atomic_skill}. This domain-aware argument instantiation ensures that each configured skill corresponds to a realizable execution on the desktop. After argument configuration, the skill  is fully specified, then ready for execution over the environment.

\subsection{Executor}

After a skill is selected and configured, the agent executes it by invoking the corresponding executable actions defined in the skill execution graph $\mathcal{G}_e$. Each execution graph specifies a parameterized realization composed of GUI interaction primitives and/or executable script actions, depending on the instantiated execution path. For execution steps that require spatial interaction with the user interface, we employ a GUI grounding model to predict interaction coordinates on the current screen. By decoupling high-level planning from low-level spatial grounding, the agent can leverage specialized perception models for accurate UI localization while allowing non-UI steps to be executed directly, improving overall execution reliability and efficiency.

Execution proceeds by traversing $\mathcal{G}_e$ in a depth-first manner to identify the next executable primitive, conditioned on the current UI state and execution context. When multiple valid successor nodes are available, the executor selects one uniformly at random by default. When edge weights are provided, the traversal policy can instead incorporate execution preferences by prioritizing successors according to their associated weights.

\subsection{Memory and Reflection.}

Our architecture incorporates a memory buffer that records previously executed skills and their observed outcomes, serving as a persistent substrate for agent reflection. For each executed skill, we generate a concise summary that captures both the skill’s intent and its resulting effect, including whether the skill succeeded, achieved the expected outcome, or failed to execute as intended.

All such summaries are stored in the memory buffer and exposed to the planner, providing an up-to-date and reflective view of the agent’s state, progress, and past decisions. Importantly, the memory module explicitly records failed skills and their contexts, enabling the planner to reason about prior mistakes, avoid repeatedly selecting ineffective actions, and adapt its strategy accordingly. Through this reflective feedback loop, the agent is encouraged to explore alternative execution paths, improving robustness and reducing unnecessary loops in long-horizon task execution.
\vspace{-1.7em}

\section{Numerical Experiment}\label{sec:experiments}

We present a comprehensive evaluation of \algacro{} and the \cuaskillagent{}. We first assess the standalone reliability of the constructed skills and their associated execution graphs across a diverse set of desktop applications. We then evaluate the effectiveness of integrating \algacro{} into an end-to-end computer-using agent on challenging real-time desktop benchmarks. Finally, we conduct ablation and robustness studies to isolate the contribution of individual components within the \algacro{} framework.

\subsection{Evaluation of Skills and Execution Graphs}
\label{subsec:atomic_skill_eval}

\paragraph{Setup.} We curate a library of 452 atomic skills spanning 17 commonly used applications on Windows OS, including File Explorer, Excel, Word, Chrome, VS Code, and system utilities, etc. To evaluate skill executability in realistic settings, we synthesize user tasks by composing atomic skills according to the skill composition graph, with arguments instantiated using the domain-aware generators described in Section~\ref{sec:atomic_skill}. Each instantiated task is executed in isolation on a virtual machine using its corresponding parameterized execution graph.
In total, we generate approximately 200K executable tasks, see examples in Appendix~\ref{appendix：synthesize_user_task_skill_composition}. We randomly sample around 1,000 tasks for evaluation. Task outcomes are assessed using GPT-5 as an LLM-based judge, with additional human screening to ensure evaluation reliability.

\begin{table*}[ht]
	\centering
	\caption{Success Rate by Application Category of \algacro{} Agent on WindowsAgentArena~\citep{bonatti2024windows}.}
	\label{tab:success_rate_by_category}
	\resizebox{1.0\linewidth}{!}{
		\begin{tabular}{lccccccc}
			\hline
			\rowcolor{gray!20}
			\textbf{Category} &
			\textbf{\# Overall} &
			\textbf{\# of Success} &
			\textbf{\# of Skills Used} &
			\textbf{\# of Distinct Skills Used} &
			\textbf{Avg Distinct Skills Per Task} &
			\textbf{Success Rate (\%)} \\
			\hline
			Chrome & 17 & 10.9 & 67  & 19 & 1.82 & 64.11 \\
			Clock & 4 & 4 & 47 & 5 & 2.50 & 100.00 \\
			File Explorer & 19 & 12 & 112 & 21 & 2.47 & {63.16} \\
			Microsoft Paint & 3 & 1 & 20 & 3 & 1.33 & 33.33 \\
			Microsoft Edge & 13 & 7 & 29 & 20 & 1.69 & 53.85 \\
			Microsoft Excel & 24 & 6 & 163 & 9 & 2.21 & 25.00 \\
			Microsoft Word & 18 & 7 & 23 & 10 & 0.94 & 38.89 \\
			Notepad & 2 & 1 & 19 & 8 & 4.00 & 50.00 \\
			Settings & 5 & 5 & 11 & 5 & 2.00 & 100.00 \\
			VLC & 21 & 11 & 56 & 16 & 1.38 & 52.38 \\
			VS Code & 24 & 10 & 71 & 11 & 1.25 & 41.67 \\
			Windows Calculator & 3 & 2 & 30 & 10 & 6.67 & 66.67 \\
			\hline
			\rowcolor{blue!8}
			\textbf{Overall} & 153 & \textbf{87.9} & 648 & \textbf{117}  & -- & \textbf{50.26} \\
			\hline
		\end{tabular}
	}
	\vspace{-1em}
\end{table*}

\paragraph{Metrics.}
We report two metrics:
\textit{(i)} Execution success rate: measures whether the synthesized tasks covering the skills successfully completes its intent,
\textit{(ii)} Average number of primitives:  counts the number of low-level actions required per skill, serving as a proxy for skill and execution complexity.
In general, a higher execution success rate indicates greater reliability of skills, while a larger average number of primitives reflects higher execution complexity

\begin{figure}
	\centering
	\includegraphics[width=\linewidth]{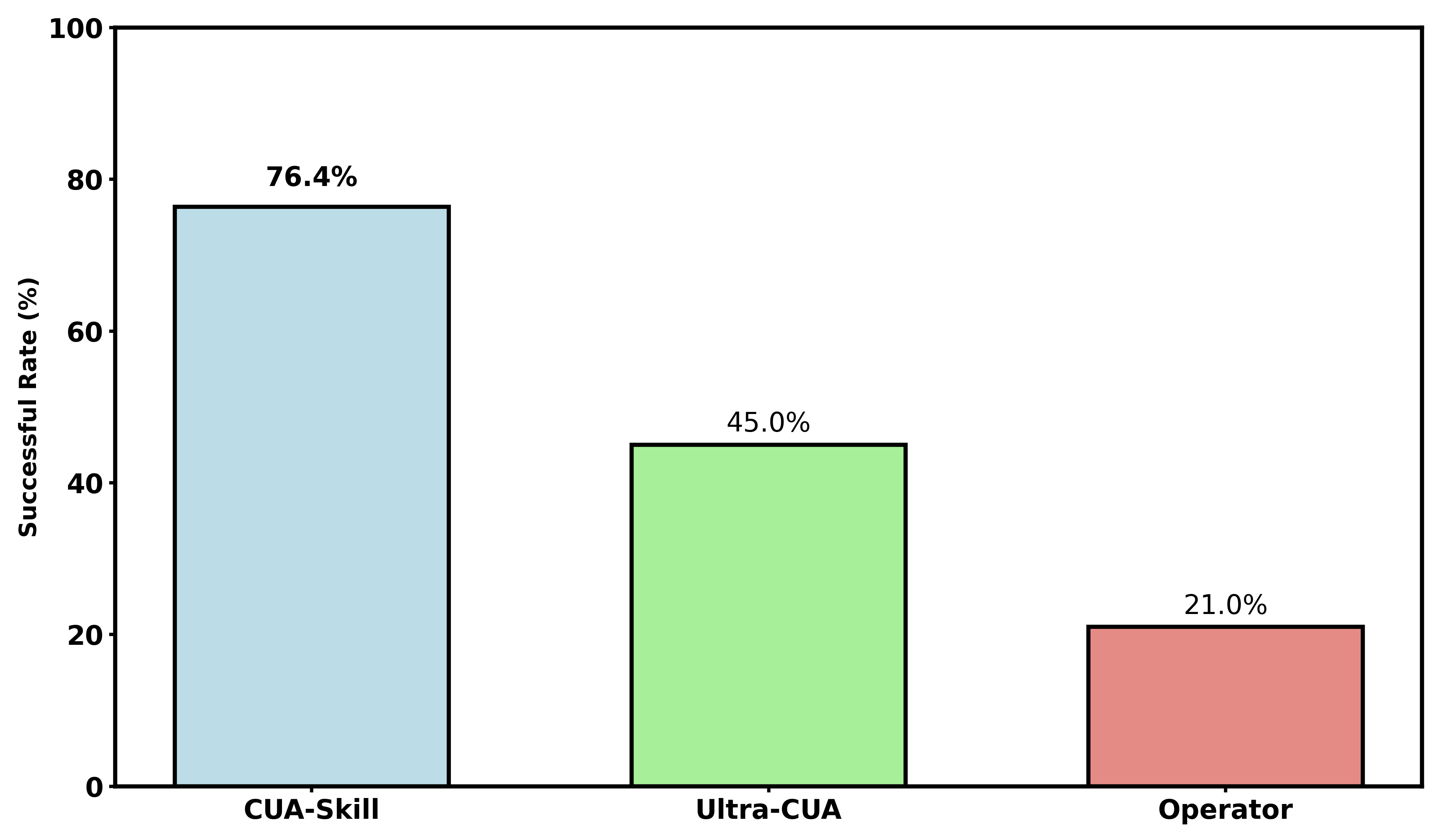}
	\caption{Synthesized User Task Successful Rate. \algacro{} is noticeablly higher than Ultra-CUA by 1.7x, and Operator by 3.6x. }
	\label{fig:trajectory_generation_comparison}
\end{figure}

\paragraph{Results.} \autoref{tab:cua_skill_graph_statistic} summarizes the execution statistics across applications. Overall, the constructed skills achieve an average success rate of \textbf{76.4\%}, with execution graphs requiring 3.75 GUI primitives per skill on average and at most 20 basic actions. This indicates strong executability while covering a broad range of interaction complexity. Applications with stable UI layouts and strong keyboard affordances (e.g., Excel, Settings, and Bing Search) exhibit higher success rates, whereas visually complex or media-heavy applications (e.g., VLC and PowerPoint) remain more challenging. These results demonstrate that \algacro{} is sufficiently reliable to serve as reusable building blocks.

\paragraph{Direct Application: CUA Trajectory Generation.}
As a direct application of skill composition, \algacro{} naturally induces executable trajectories by composing skills through their parameterized execution and composition graphs. This process yields complete, low-cost, and high-success trajectories that can serve a variety of downstream purposes. We compare \algacro{} with existing trajectory generation systems, including UltraCUA~\citep{yang2025ultracua}, which reports a success rate of 45\%, as well as OpenAI Operator~\citep{openai2025operator} evaluated on the same synthesized user tasks. As shown in \autoref{fig:trajectory_generation_comparison}, trajectories generated by \algacro{} achieve substantially higher success rates \textcolor{ForestGreen}{\textbf{1.7$\times$}}-\textcolor{ForestGreen}{\textbf{3.6$\times$}} higher than UltraCUA and Operator, respectively. These results suggest that \algacro{} can alleviate the training data scarcity bottleneck of CUA.

\subsection{End-to-End Performance of the \cuaskillagent}
\label{subsec:e2e_agent_eval}

We next evaluate the effectiveness of \cuaskillagent{} in an end-to-end computer-using agent setting, where the agent operates directly from natural-language user instructions without skill composition available. Unlike the synthesized skill compositions in Section~\ref{subsec:atomic_skill_eval}, \cuaskillagent{} must autonomously decide which skill to invoke, when to invoke it, and how to configure its arguments based on the current UI state and execution history. Consequently, failures may stem not only from execution errors, but also from imperfect skill retrieval, mis-ranking, or incorrect argument instantiation. This evaluation therefore provides a stringent test of whether structured and parameterized skill abstractions can support robust decision making in realistic user tasks.

\begin{table}[h]
	\centering
	\caption{Performance on the WAA Benchmark.}
	\label{tab:waa_benchmark}
	\resizebox{1.0\linewidth}{!}{
		\begin{tabular}{l c c}
			\hline
			\rowcolor{gray!20}\textbf{System} & \textbf{Success Rate (\%)} & \textbf{\# of Steps} \\
			\hline
			Human Performance~\citep{bonatti2024windows} & 74.5 & --\\
			\midrule
			NAVI~\citep{bonatti2024windows} (GPT4o) & 19.5 & 15 \\
			UI-TARS1~\citep{qin2025ui} (Qwen2VL-FT) & 15.7 & 50 \\
			UITARS-1.5-7B~\citep{qin2025ui} (Qwen2.5VL-FT) & 18.1 & 15 \\
			STEVE~\citep{lu2025steve} (GPT4o) & 20.1 & 40 \\
			Agent-S~\citep{agashe2024agents} (GPT-4o) & 18.2 & 15 \\
			Agent-S2~\citep{agashe2025agents2} (Claude-3.5-Sonnet) & 29.8 & 15\\
			UFO-2~\citep{zhang2025ufo2} (GPT-o1) & 30.5 & 30\\
			Ultra-CUA-7B~\citep{yang2025ultracua} (QWen2.5VL-FT) & 21.1 & 15\\
			AgentS3~\citep{gonzalez2025unreasonable} (GPT-5) & 49.0 & 50 \\
			AgentS3~\citep{gonzalez2025unreasonable} (GPT-5) (Best-of-3) & 56.6 & 100 \\
			Operator~\citep{openai2025operator}  & 37.4 & 50 \\
			\midrule
			\rowcolor{blue!8}\textbf{\algacro{} Agent} (GPT-5) & \textbf{50.3} & 30 \\ 
			\rowcolor{blue!8}\textbf{\algacro{} Agent} (GPT-5, Best of 3) & \textbf{57.5} & 30 \\ 
			\hline
		\end{tabular}
	}
	\vspace{-1em}
\end{table}

\paragraph{Benchmarks and Metrics.} Since \algacro{} primarily focus on Windows OS, we naturally evaluate \cuaskillagent{} on the popular WindowsAgentArena benchmark~\citep{bonatti2024windows}. We report success rate and the number of (distinct) skills used per task. The former one indicates the overall performance of the agent that we built. The later two indicates the coverage of the \algacro{}. 
\vspace{-1em}

\paragraph{Results.} \autoref{tab:success_rate_by_category} reports per-application success rates of the \cuaskillagent{}. Averaged across all evaluated tasks, the agent achieves a 50.26\% success rate while using an average of 2.22 distinct skills per task. Performance varies across applications: system utilities and configuration tasks are solved reliably, whereas document editing and spreadsheet workflows remain more challenging due to dense UI interactions. These results demonstrate the effectiveness of structured skill reuse, while highlighting remaining challenges in complex application workflows. We further compare \cuaskillagent{} with existing CUA systems in \autoref{tab:waa_benchmark}. With GPT-5 as the planner, \cuaskillagent{} achieves a state-of-the-art best-of-three success rate of 57.5\%, significantly outperforming existing approaches by a large margin. Moreover, in addition to its strong performance, \cuaskillagent{} completes tasks efficiently, requiring at most 30 execution steps. Notably, across all WAA evaluations, the agent invokes only 117 distinct skills out of the 478 available in \algacro{}, indicating that the performance gains arise from general-purpose skill abstractions rather than benchmark-specific engineering.
\vspace{-1em}

\subsection{Ablation Study}\label{subsec:ablation_atomic}

We studied the impact of different planners and the gain of \algacro{} to improve computer use performance.

\begin{wraptable}{r}{0.5\linewidth}
	\centering
	\vspace{-0.5em}
	\hspace{-2em}
	\caption{LLM Backbones.}
	\label{tab:model_comparison}
	\scriptsize
	\resizebox{1.0\linewidth}{!}{
		\begin{tabular}{lc}
			\toprule
			\textbf{Model Configuration} & \textbf{SR (\%)} \\ 
			\midrule
			Qwen3-VL-32B-Instruct         & \cellcolor{yellow!10}{11.77} \\
			GPT-4o                       & \cellcolor{yellow!20}{28.10} \\
			GPT-5 (Minimal Reasoning)    & \cellcolor{orange!10}{33.31} \\
			GPT-5 (Low Reasoning)        & \cellcolor{orange!20}{50.26} \\
			\bottomrule
		\end{tabular}
	}
	\vspace{-1.7em}
\end{wraptable}
\textbf{LLM Planner.} Skills are designed to be model-agnostic and compatible with a wide range of LLM backbones. As shown in \autoref{tab:model_comparison}, we evaluate \cuaskillagent{} using Qwen3-VL-32B-Instruct, GPT-4o, and GPT-5. The results show a clear positive correlation between agent performance and the capability of the underlying language model, with GPT-5 achieving  higher successful rate than less capable backbones. We further ablate the effect of reasoning depth within GPT-5. Increasing the reasoning level consistently improves task success, rising from 33.31\% under minimal reasoning to 50.26\% under low reasoning. This trend indicates that stronger reasoning benefits the usage of \algacro{} for computer use.
\vspace{-0.2em}

\begin{table}[ht]
	\centering
	\caption{Skill Integration Across Different Backbones on WAA.}
	\label{tab:skill_saliency}
	\resizebox{1.0\linewidth}{!}{
		\begin{tabular}{@{}lccc@{}}
			\toprule
			\textbf{Model Backbone} & \textbf{Baseline (No Skill)} & \textbf{With Skills} & \textbf{Improvement ($\Delta$)} \\ \midrule
			Qwen3-VL-32B-Instruct   & 6.54\% & 11.77\%  & \cellcolor{gray!20}{\textcolor{green!60!black}{\textbf{+5.23\% ($\uparrow$)}}}  \\
			GPT-4o & 19.60\% & 28.10 & \cellcolor{gray!20}{\textcolor{green!60!black}{\textbf{+8.50\% ($\uparrow$)}}} \\
			GPT-5 & 34.64\% & 50.26\%  & \cellcolor{gray!20}{\textcolor{green!60!black}{\textbf{+15.62\% ($\uparrow$)}}} \\ 
			\bottomrule
		\end{tabular}
	}
	\vspace{-1.7em}
\end{table}

\paragraph{Saliency of Skills.}
\autoref{tab:skill_saliency} demonstrates that skill augmentation consistently improves agent performance across all evaluated LLM backbones, with gains scaling alongside model capability. For Qwen3-VL-32B-Instruct, skills deliver a substantial improvement (\textcolor{green!60!black}{\textbf{+5.23\%}}). GPT-4o exhibits a larger gain (\textcolor{green!60!black}{\textbf{+8.50\%}}), reflecting improved reliability in skill selection and configuration. For GPT-5, skill integration yields the largest improvement (\textcolor{green!60!black}{\textbf{+15.62\%}}). 
\vspace{-0.8em}

\section{Conclusion}
We present \algacro{} and \cuaskillagent{}, a skill-centric framework that encodes human computer-use knowledge as reusable, parameterized skills with execution and composition graphs. \algacro{} is highly transferable across tasks and applications, and directly enables high-success executable trajectory generation. Evaluations on WindowsAgentArena show consistent performance gains across LLM backbones, establishing \algacro{} as a practical, model-agnostic foundation for scalable desktop agents.

\bibliography{reference}
\bibliographystyle{plain}

\clearpage
\onecolumn
\newpage
\appendix

\section{Author List}\label{appendix:authorlist}
\vspace{-0.5em}

\textbf{Project Lead:} Tianyi Chen, \href{mailto:Tianyi.Chen@microsoft.com}{Tianyi.Chen@microsoft.com}.\\
\textbf{Primary Contributor:} Tianyi Chen, Yinheng Li, Michael Solodko, Sen Wang, Nan Jiang, Tingyuan Cui.\\
\textbf{Contributor:} Junheng Hao, Jongwoo Ko, Sara Abdali, Qing Xiao.\\
\textbf{Leadership:} Justin Wagle, Pashmina Cameron, Suzhen Zheng, Leon Xu, Hao Fan, Kazuhito Koishida.
\vspace{-1em}

\section{More Related Works}\label{appendix:more_related_works}
\vspace{-0.5em}

\paragraph{GUI Grounding in Desktop Agents}

GUI grounding is central to dekstop-agent planning since each natural-language step must be bound to a specific on-screen target before an action can be executed ~\citep{hui-etal-2025-winspot}. In desktop environments, grounding must be repeated after every state change ~\citep{xie2024osworld}. OS-agent evaluations show that minor grounding errors quickly compound into multi-step failures ~\citep{bonatti2024windows}. Recent work improves this by strengthening UI perception and reference resolution ~\citep{zhang2025phiground, wang2025uitars2,zhao2025robustness}. \algacro{} uses grounding as a planning constraint, aligning each candidate action with spatial and contextual cues from the current interface state.
\vspace{-0.7em}

\paragraph{Retrieval-Augmented Planning}
Retrieval-augmented planning interleaves planning and action, letting an agent revise its next step using environment feedback instead of committing to a full plan up front ~\citep{ac1f09077393404a8bea5141d8710259}. A challenge is tool orchestration. Agents must decide whether an external tool is needed and select an appropriate tool given the current objective and context ~\citep{huang2023metatool}. Retrieval addresses scalability by narrowing a large action/tool space to a small set of relevant candidates. These candidates can then be ranked and invoked within the reasoning loop rather than treated as static menus ~\citep{braunschweiler-etal-2025-toolreagt, qu2025tool, schick2023toolformer}. \algacro{} follows this paradigm by retrieving and ranking atomic template conditions on the current goal, enabling more targeted planning and tool selection.
\vspace{-0.4em}

\section{Example of Ensembled Query Generation}\label{appendix:ensemble_query_generation}

\begin{tcolorbox}[
	colback=green!10!white,    
	colframe=green!60!black,   
	title=\textbf{Ensembled Query Generation Example},
	boxrule=0.4pt,           
	arc=2pt,                 
	left=4pt, right=4pt, top=2pt, bottom=2pt
	]
	\textbf{Instruction}:  Next: Open Edge Home Page. \\
	\textbf{Query 1}: {Open home page in Edge.} \\
	\textbf{Query 2}: {Double-click Microsoft Edge icon to open it and navigate to the home page.} \\
	\textbf{Query 3}: {Use Windows menu to launch Edge.}
\end{tcolorbox}
\vspace{-0.3em}

\section{Example of Synthesized Tasks by Skill Composition Graph}\label{appendix：synthesize_user_task_skill_composition}

\begin{tcolorbox}[
	colback=yellow!10!white,    
	colframe=yellow!60!black,   
	title=\textbf{Synthesized User Task Example 1 upon Skill Composition Graph},
	boxrule=0.4pt,           
	arc=2pt,                 
	left=4pt, right=4pt, top=2pt, bottom=2pt
	]
	Domain: Excel\\
	Instruction: Open the `betawacc.xlsx' file, rename the sheet1 as company analysis and fulfill the average column.\\
	Steps: (A sequence of skills to complete the instruction)\\
	\textbf{ExcelOpenExistingWorkbook}, \texttt{file\_path=betawacc.xlsx}.\\
	\textbf{ExcelRenameSheet}, \texttt{target\_sheet\_name=sheet1}, \texttt{new\_sheet\_name=company analysis}.\\
	\textbf{ExcelInsertFunctionCall}, \texttt{target\_cell=F7}, \texttt{function\_call\_command=AVERAGE(C7:E7)}.\\
	\textbf{ExcelAutoFillDown}, \texttt{start\_cell=F7}, \texttt{end\_cell=F10}.
\end{tcolorbox}
\vspace{-0.2em}
\begin{tcolorbox}[
	colback=yellow!10!white,    
	colframe=yellow!60!black,   
	title=\textbf{Synthesized User Task Example 2 upon Skill Composition Graph},
	boxrule=0.4pt,           
	arc=2pt,                 
	left=4pt, right=4pt, top=2pt, bottom=2pt
	]
	Domain: Calculator\\
	Instruction: Calculate $398-174\times\sqrt{505}$\\
	Steps: (A sequence of skills to complete the instruction)\\
	\textbf{CalculatorLaunch}.\\
	\textbf{CalculatorSwitchMode}, \texttt{mode\_name=scientific}.\\
	\textbf{CalculatorEnterNumber}, \texttt{number=398}.\\
	\textbf{CalculatorSubtract}.\\
	\textbf{CalculatorEnterNumber}, \texttt{number=174}.\\
	\textbf{CalculatorMultiply}.\\
	\textbf{CalculatorSquareRoot}, \texttt{number=505}.\\
	\textbf{CalculatorEquals}.
\end{tcolorbox}

\newpage
\section{Case Study}\label{appendix:case_study}
\begin{figure}[H]
\vspace{-0.5em}
\centering
\begin{minipage}{0.9\linewidth}
\paragraph{Case Study: Skill: \texttt{ClockCreateTimer}.}
\small
\textbf{Task:} Create a 25 minute timer called Pomodoro Session.
\vspace{0.5em}
\hrule
\vspace{0.5em}

\begin{minipage}{0.32\linewidth}
    \centering
    \includegraphics[width=\linewidth]{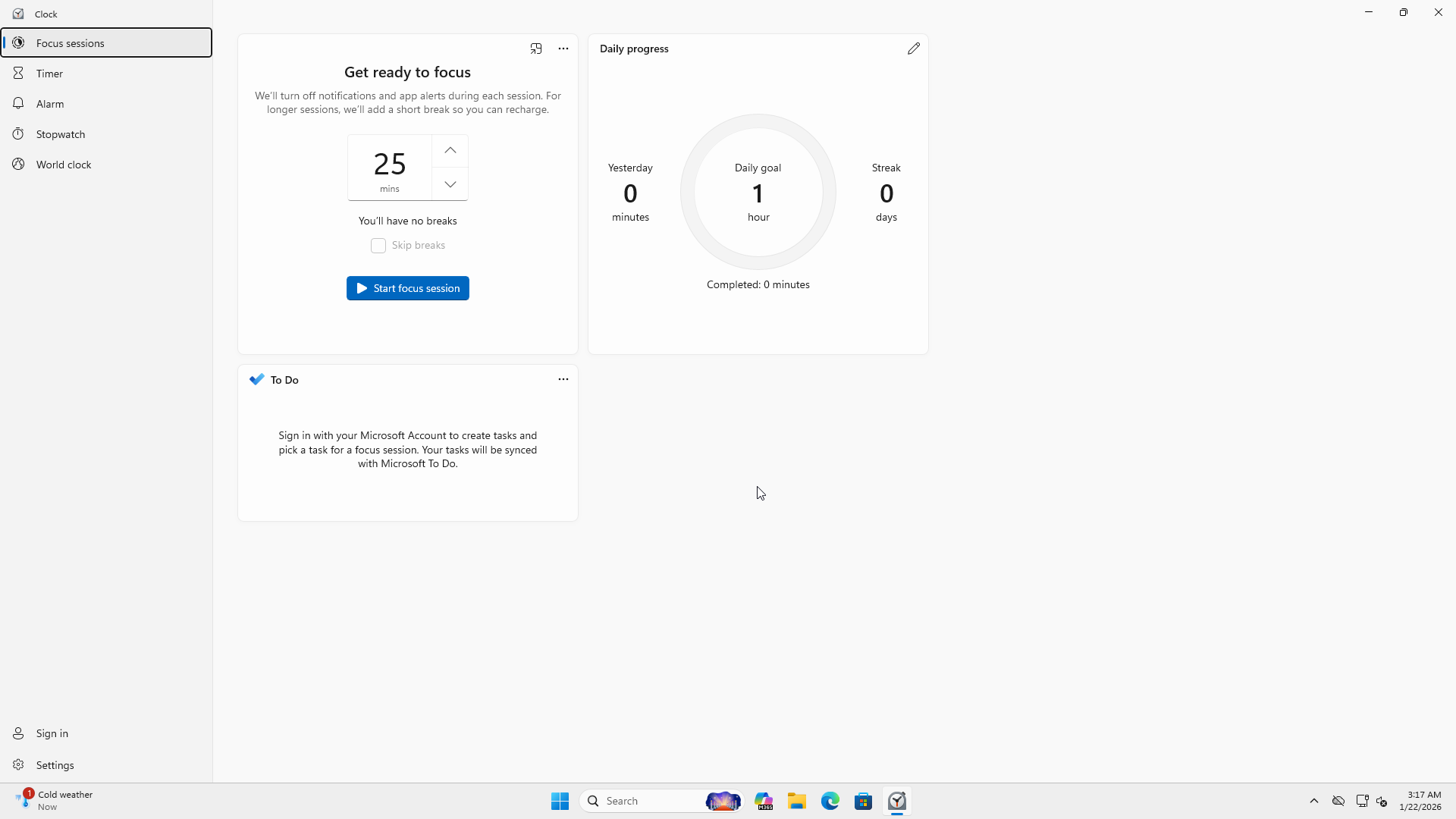}
\end{minipage}
\hfill
\begin{minipage}{0.58\linewidth}
\small
\textbf{Step 1: ClockSwitchTab}\\
\textbf{Reasoning.} Click on the `Timer' tab to switch view.\\
\textbf{Argument Instantiation.}
\begin{itemize}[leftmargin=1em]
    \item \texttt{tab}: Timer.
\end{itemize}
\end{minipage}

\begin{minipage}{0.32\linewidth}
    \centering
    \includegraphics[width=\linewidth]{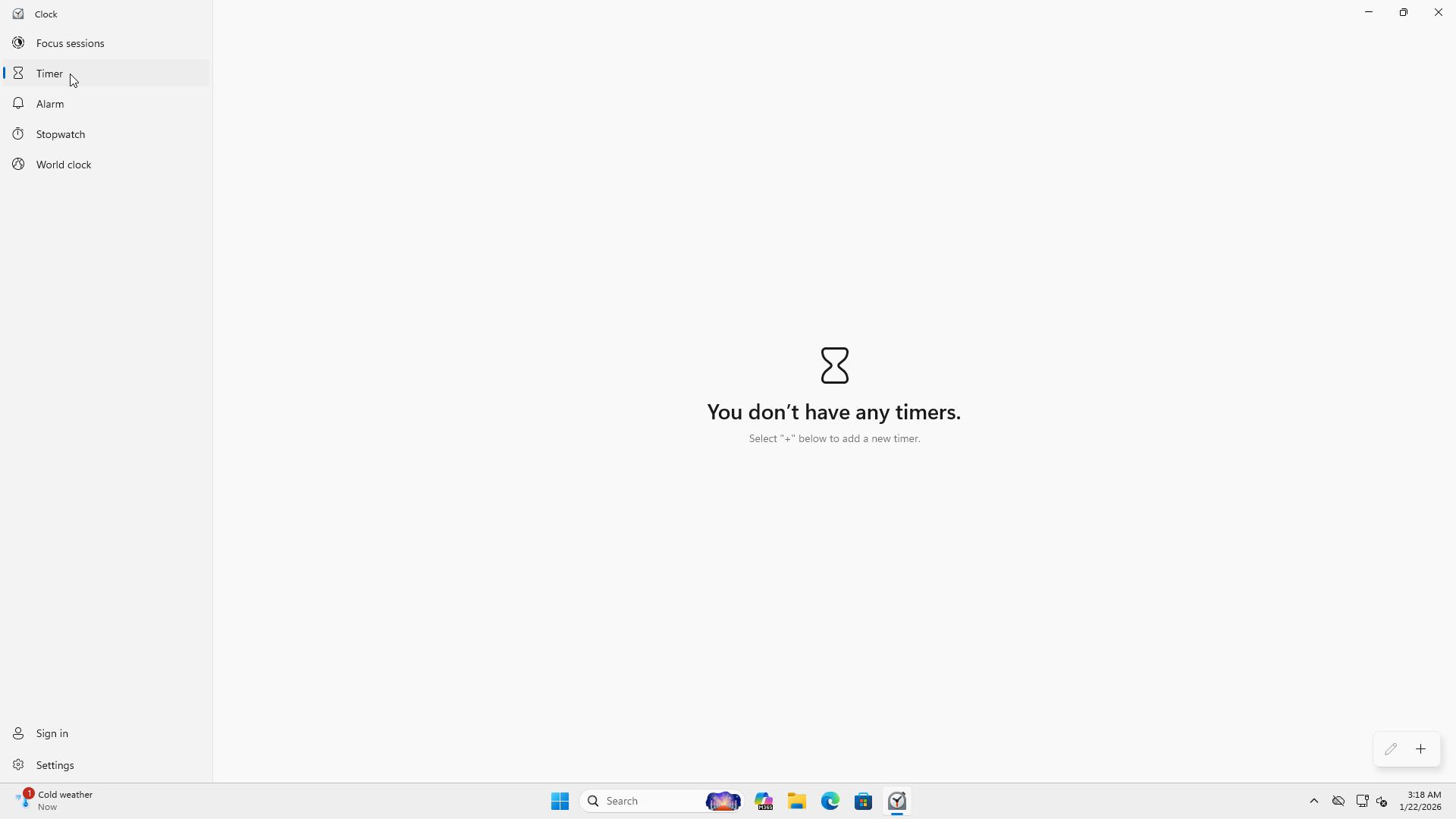}
\end{minipage}
\hfill
\begin{minipage}{0.58\linewidth}
\small
\textbf{Step 2: \texttt{SingleClickAction}}\\
\textbf{Reasoning.}
Click add timer button.\\
\textbf{Argument Instantiation.}
\begin{itemize}[leftmargin=1em]
    \item \texttt{Coordinate:} Call grounding model.
    \item \texttt{Button:} Left
\end{itemize}
\end{minipage}

\begin{minipage}{0.32\linewidth}
    \centering
    \includegraphics[width=\linewidth]{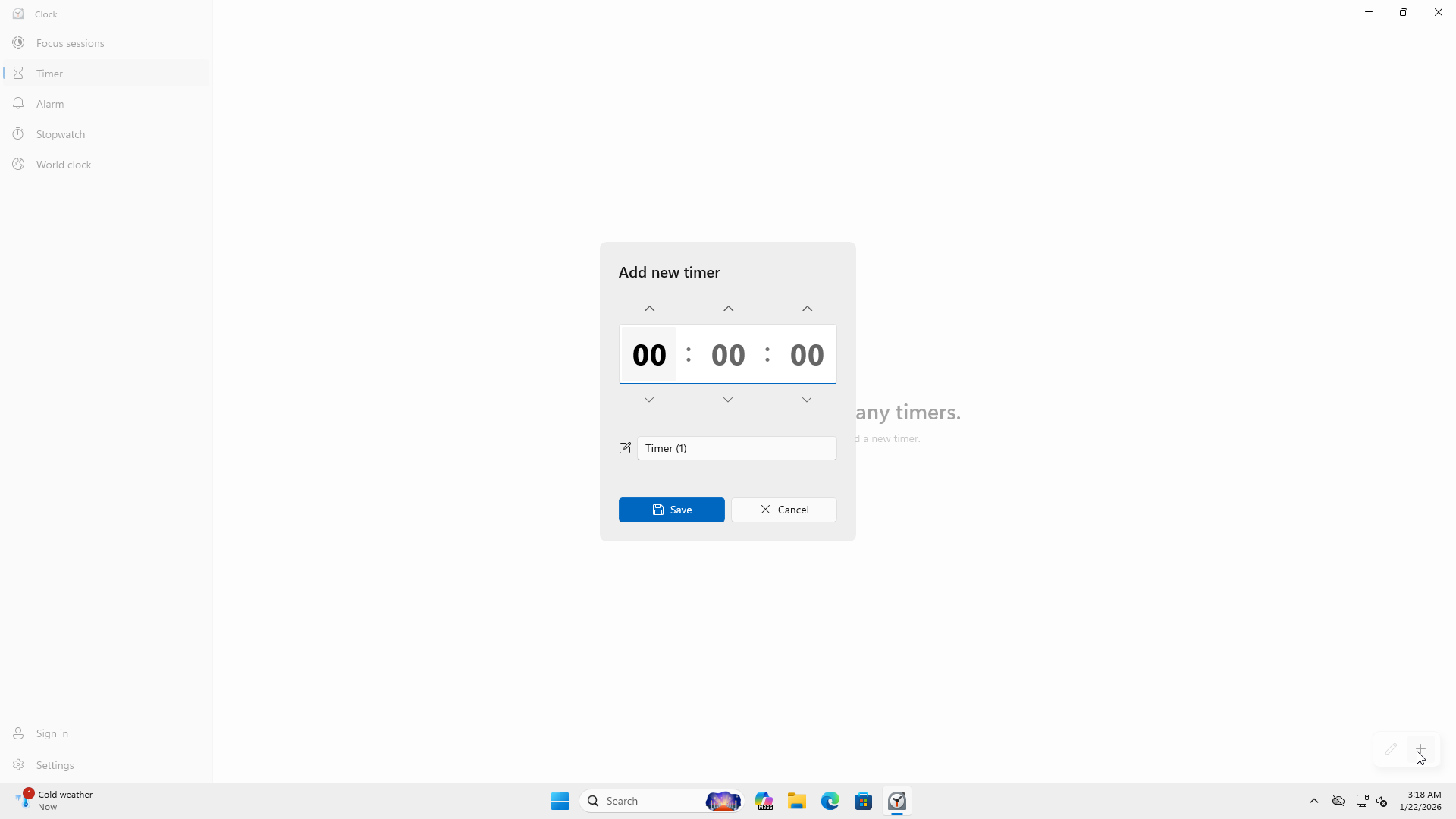}
\end{minipage}
\hfill
\begin{minipage}{0.58\linewidth}
\small
\textbf{Step 3: \texttt{SingleClickAction}}\\
\textbf{Reasoning.}
Focus on minutes input.\\
\textbf{Argument Instantiation.}
\begin{itemize}[leftmargin=1em]
    \item \texttt{Coordinate:} Call grounding model.
    \item \texttt{Button:} Left
\end{itemize}
\end{minipage}

\begin{minipage}{0.32\linewidth}
    \centering
    \includegraphics[width=\linewidth]{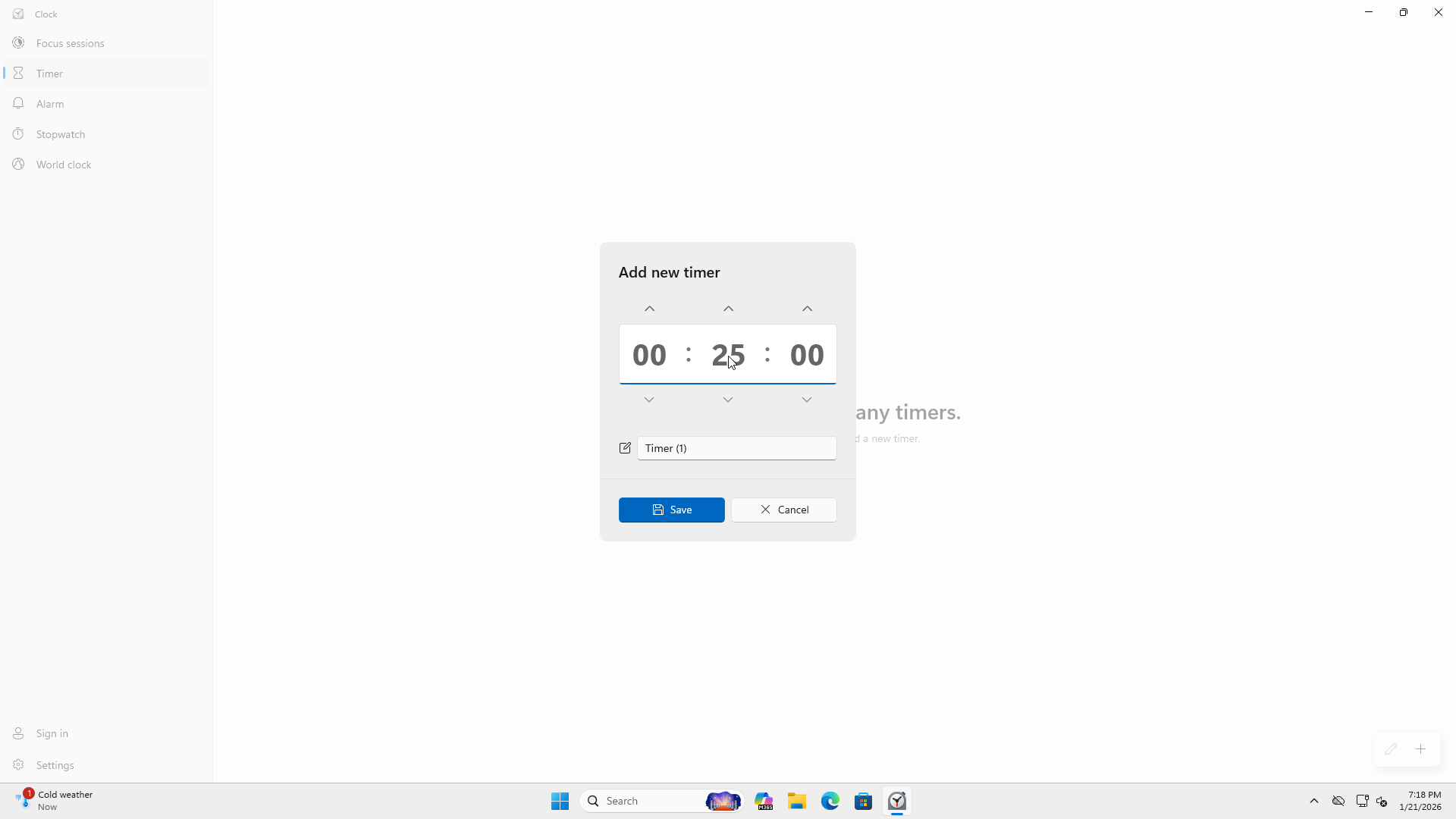}
\end{minipage}
\hfill
\begin{minipage}{0.58\linewidth}
\small
\textbf{Step 4: \texttt{TypeAction}}\\
\textbf{Reasoning.}
Enter minutes `25'.\\
\textbf{Argument Instantiation.}
\begin{itemize}[leftmargin=1em]
    \item \texttt{Input\_mode:} `keyboard'
    \item \texttt{Text:} 25
\end{itemize}
\end{minipage}

\begin{minipage}{0.32\linewidth}
    \centering
    \includegraphics[width=\linewidth]{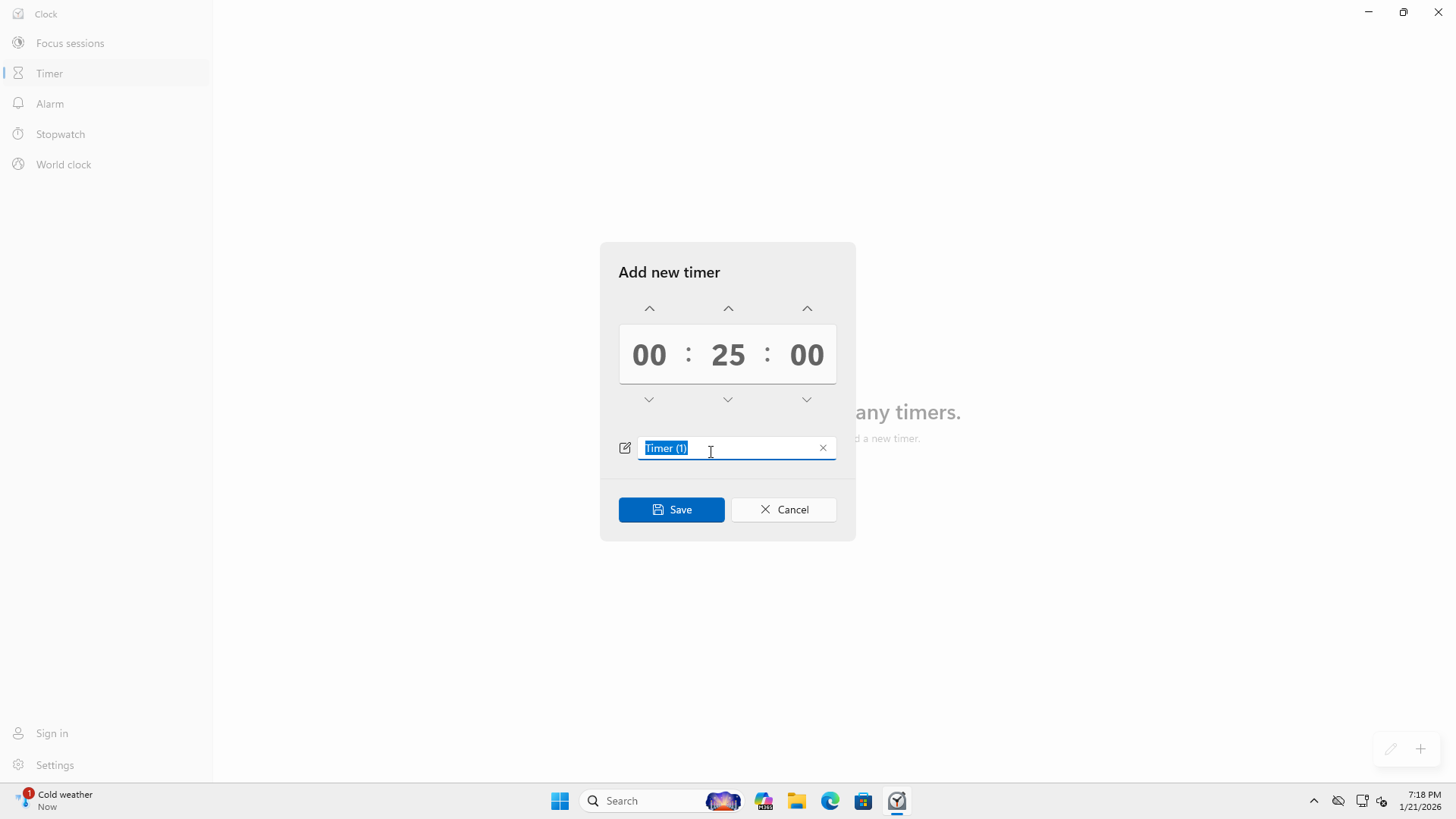}
\end{minipage}
\hfill
\begin{minipage}{0.58\linewidth}
\small
\textbf{Step 5: \texttt{SingleClickAction}}\\
\textbf{Reasoning.} Locate and click the timer name input field (e.g., placeholder `Name' or `Timer name').\\
\textbf{Argument Instantiation.}
\begin{itemize}[leftmargin=1em]
    \item \texttt{Coordinate:} Call grounding model.
    \item \texttt{Button:} Left.
\end{itemize}
\end{minipage}

\begin{minipage}{0.32\linewidth}
    \centering
    \includegraphics[width=\linewidth]{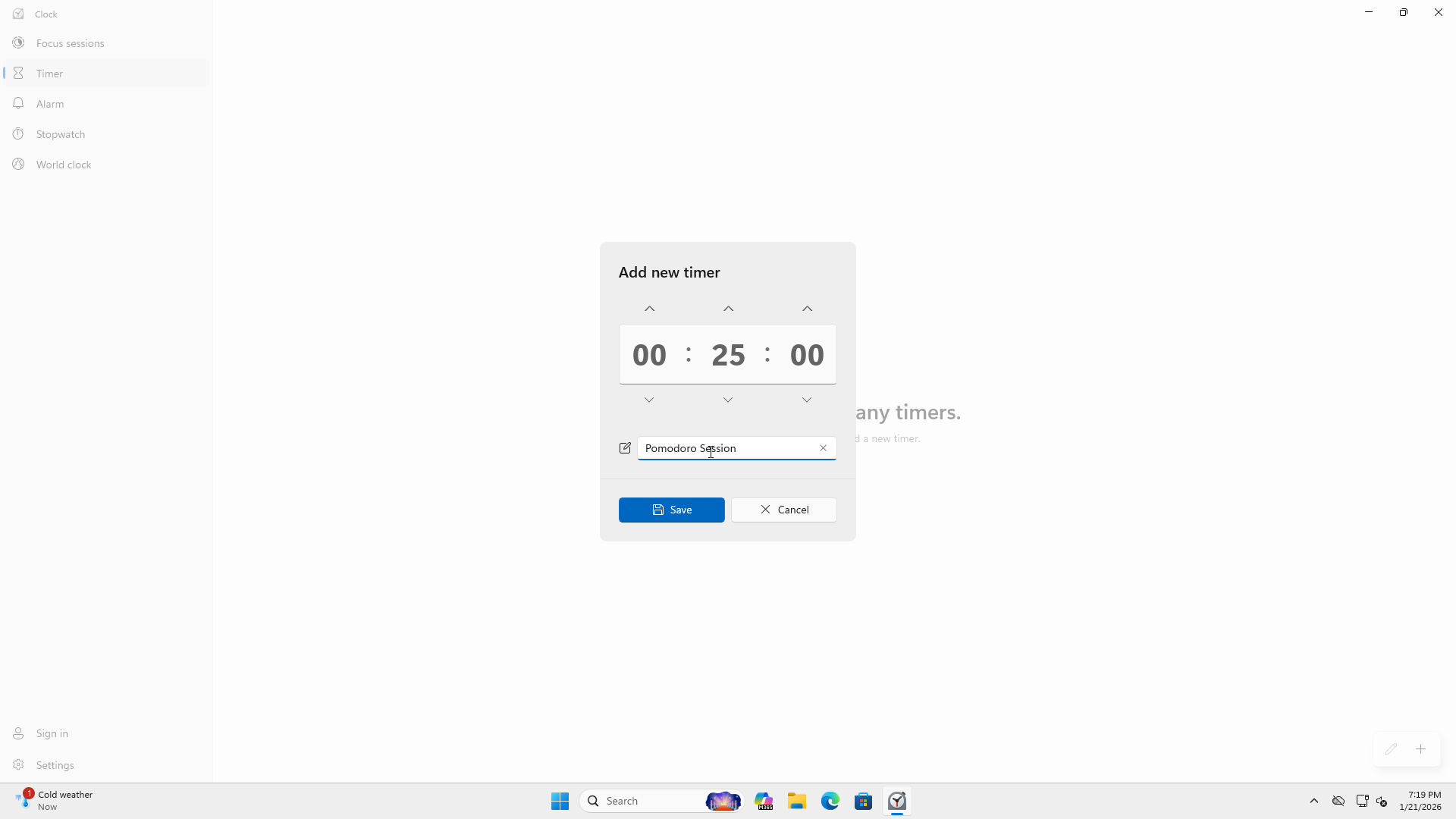}
\end{minipage}
\hfill
\begin{minipage}{0.58\linewidth}
\small
\textbf{Step 6:  \texttt{TypeAction}}\\
\textbf{Reasoning.} Type timer label `Pomodoro Session'.\\
\textbf{Argument Instantiation.}
\begin{itemize}[leftmargin=1em]
    \item \texttt{Input\_mode:} `copy\_paste'
    \item \texttt{Text:} `Pomodoro Session'.
\end{itemize}
\end{minipage}

\begin{minipage}{0.32\linewidth}
    \centering
    \includegraphics[width=\linewidth]{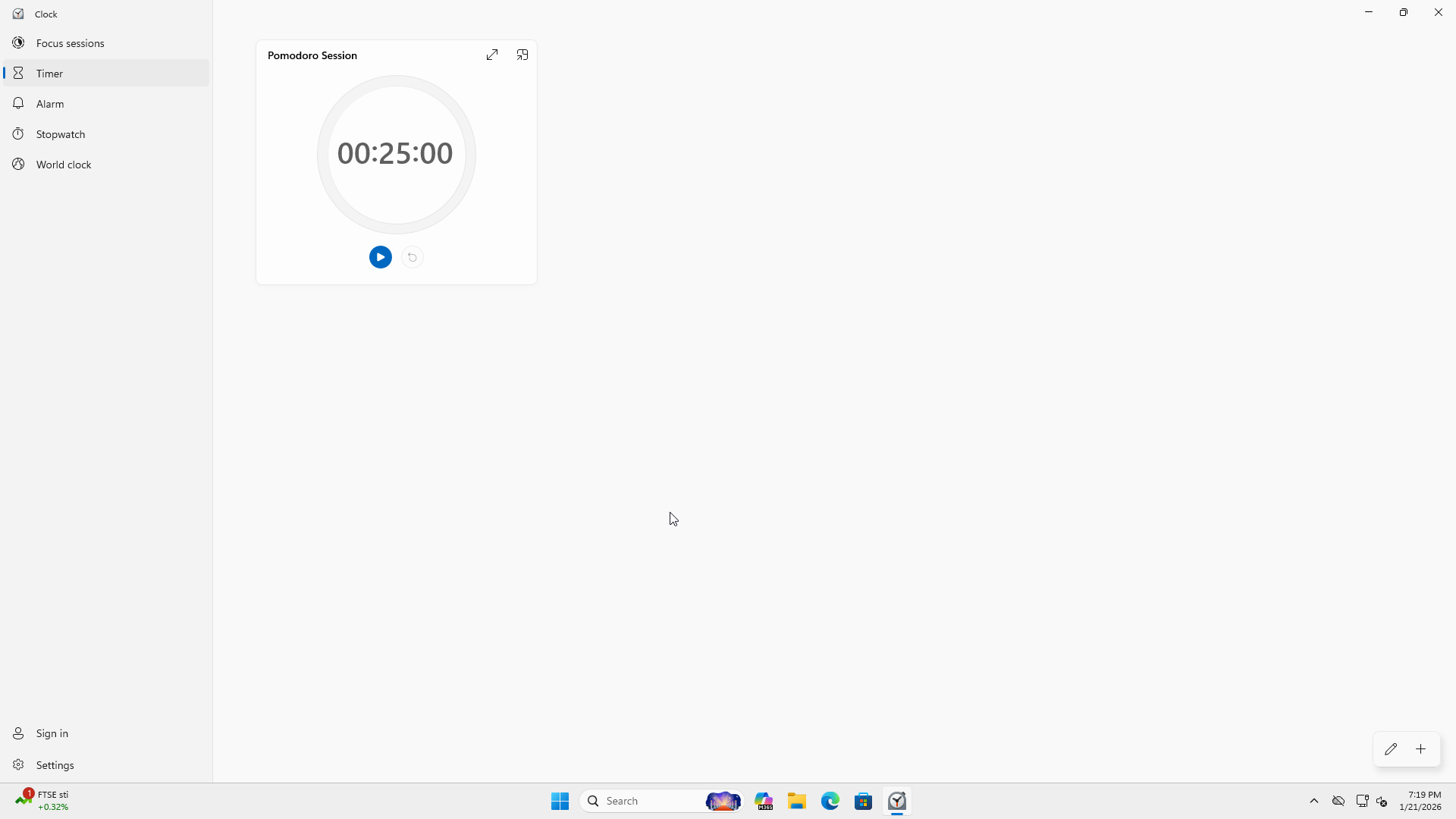}
\end{minipage}
\hfill
\begin{minipage}{0.58\linewidth}
\small
\textbf{Step 7:  \texttt{SingleClickAction}}\\
\textbf{Reasoning.} Save timer..\\
\textbf{Argument Instantiation.}
\begin{itemize}[leftmargin=1em]
    \item \texttt{Coordinate:} Call grounding model.
    \item \texttt{Button:} Left.
\end{itemize}
\end{minipage}

\hrule
\vspace{0.4em}

\small
\textbf{Outcome.} The agent successfully completed the skill.

\textbf{Key Insight.} Skill arguments either need Planner to configure or call grounding model to predict, e.g., the coordinate.
\end{minipage}
\vspace{-1em}
\end{figure}

\begin{figure}[H]
\centering
\begin{minipage}{0.95\linewidth}
\paragraph{Case Study: Skill: \texttt{FileExplorerCreateNewFolder}.}
\small

\textbf{Task:} Create a new folder named Logs inside Downloads.
\vspace{0.4em}
\hrule
\vspace{0.4em}

\begin{minipage}{0.32\linewidth}
    \centering
    \includegraphics[width=\linewidth]{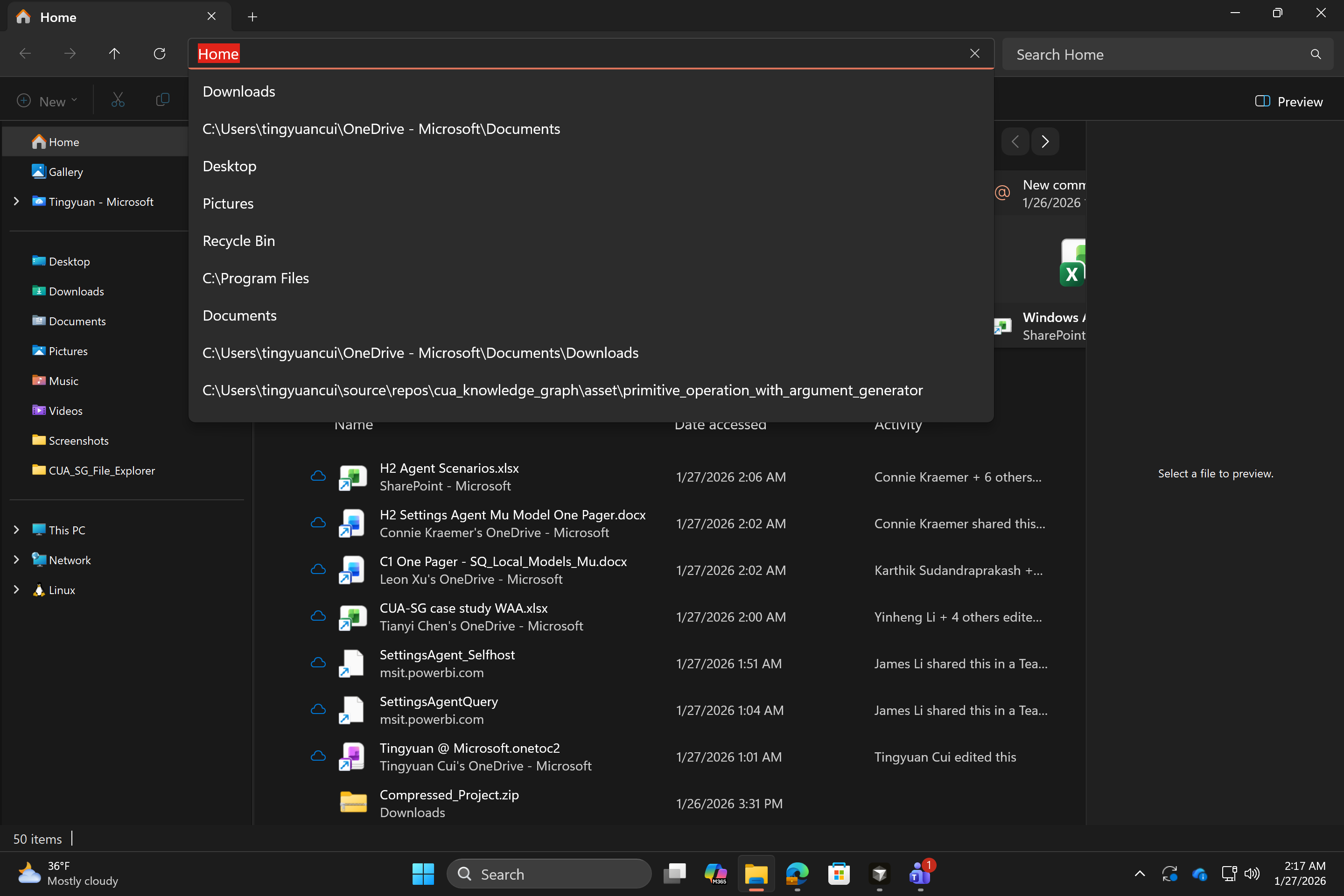}
\end{minipage}
\hfill
\begin{minipage}{0.58\linewidth}
\small
\textbf{Step 1: \texttt{HotKeyAction}}\\
\textbf{Reasoning.} Focus the address bar.\\
\textbf{Argument Instantiation.}
\begin{itemize}[leftmargin=1em]
    \item \texttt{keys}: [`ctrl', `l'].
\end{itemize}
\end{minipage}

\vspace{0.4em}

\begin{minipage}{0.32\linewidth}
    \centering
    \includegraphics[width=\linewidth]{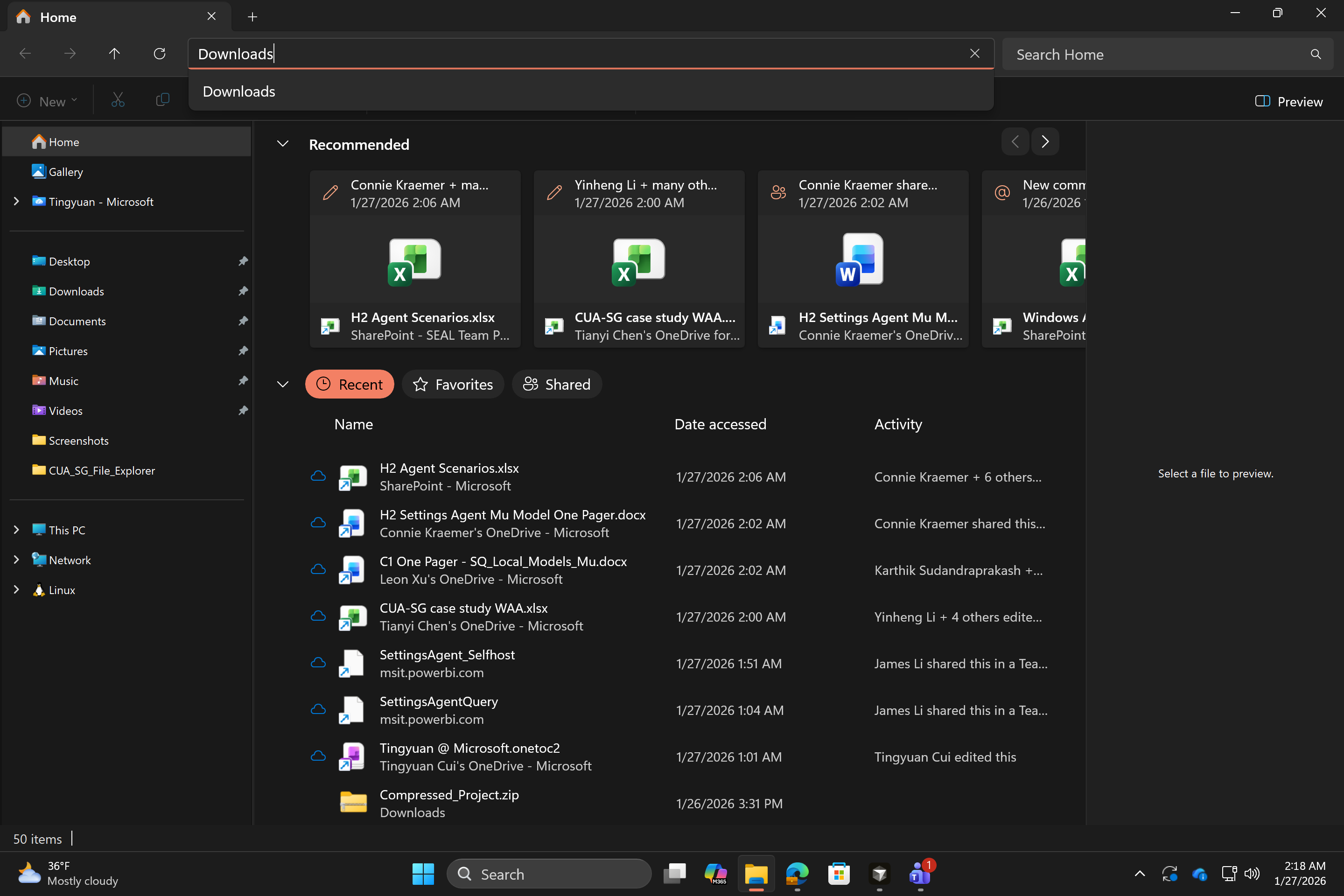}
\end{minipage}
\hfill
\begin{minipage}{0.58\linewidth}
\small
\textbf{Step 2: \texttt{TypeAction}}\\
\textbf{Reasoning.} Type path `Downloads` to search for it.\\
\textbf{Argument Instantiation.}
\begin{itemize}[leftmargin=1em]
    \item \texttt{input\_mode:} `keyboard'
    \item \texttt{text:} `Downloads'
\end{itemize}
\end{minipage}

\vspace{0.4em}

\begin{minipage}{0.32\linewidth}
    \centering
    \includegraphics[width=\linewidth]{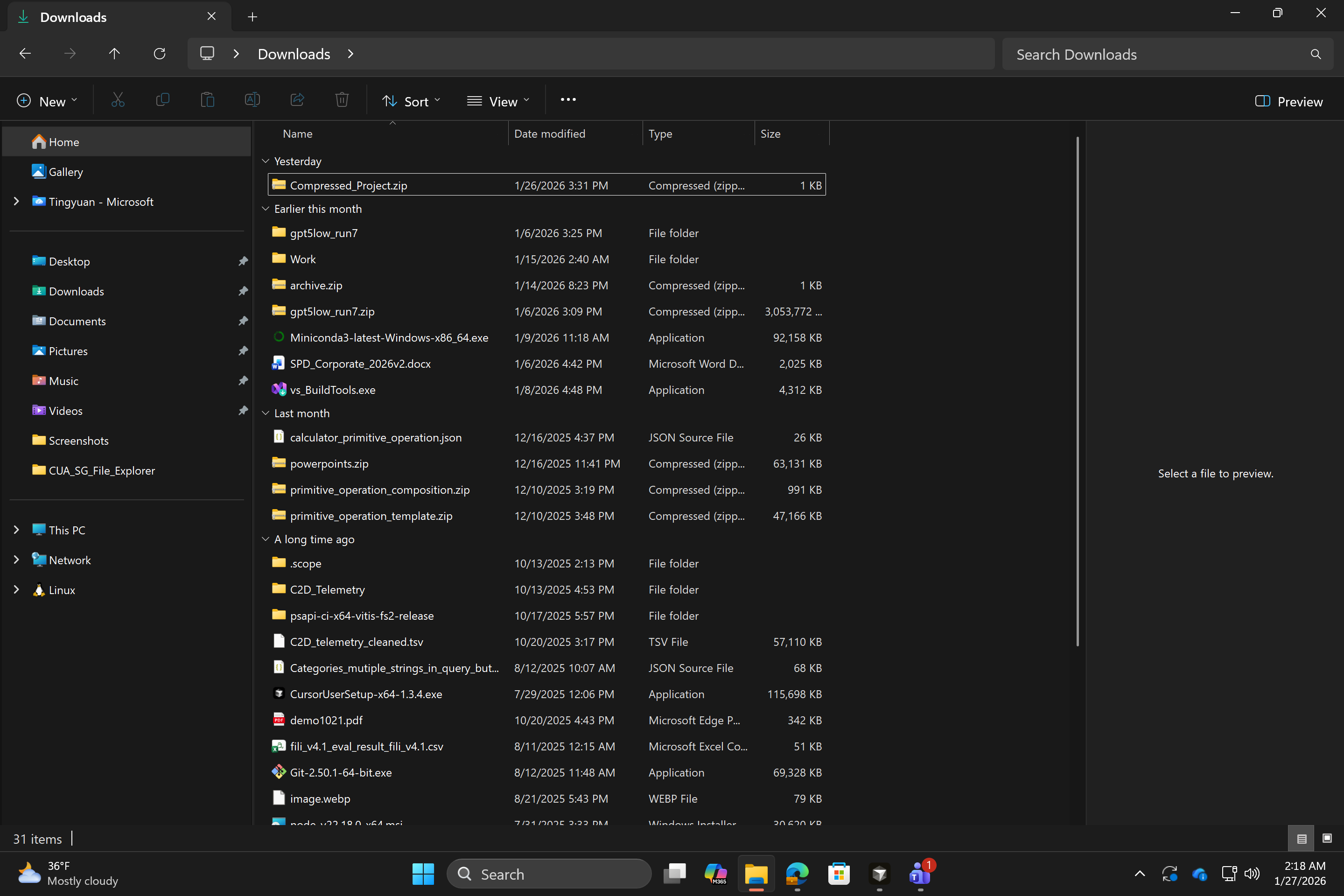}
\end{minipage}
\hfill
\begin{minipage}{0.58\linewidth}
\small
\textbf{Step 3: \texttt{HotKeyAction}}\\
\textbf{Reasoning.} Open folder.\\
\textbf{Argument Instantiation.}
\begin{itemize}[leftmargin=1em]
    \item \texttt{key:} `Enter'
\end{itemize}
\end{minipage}

\begin{minipage}{0.32\linewidth}
    \centering
    \includegraphics[width=\linewidth]{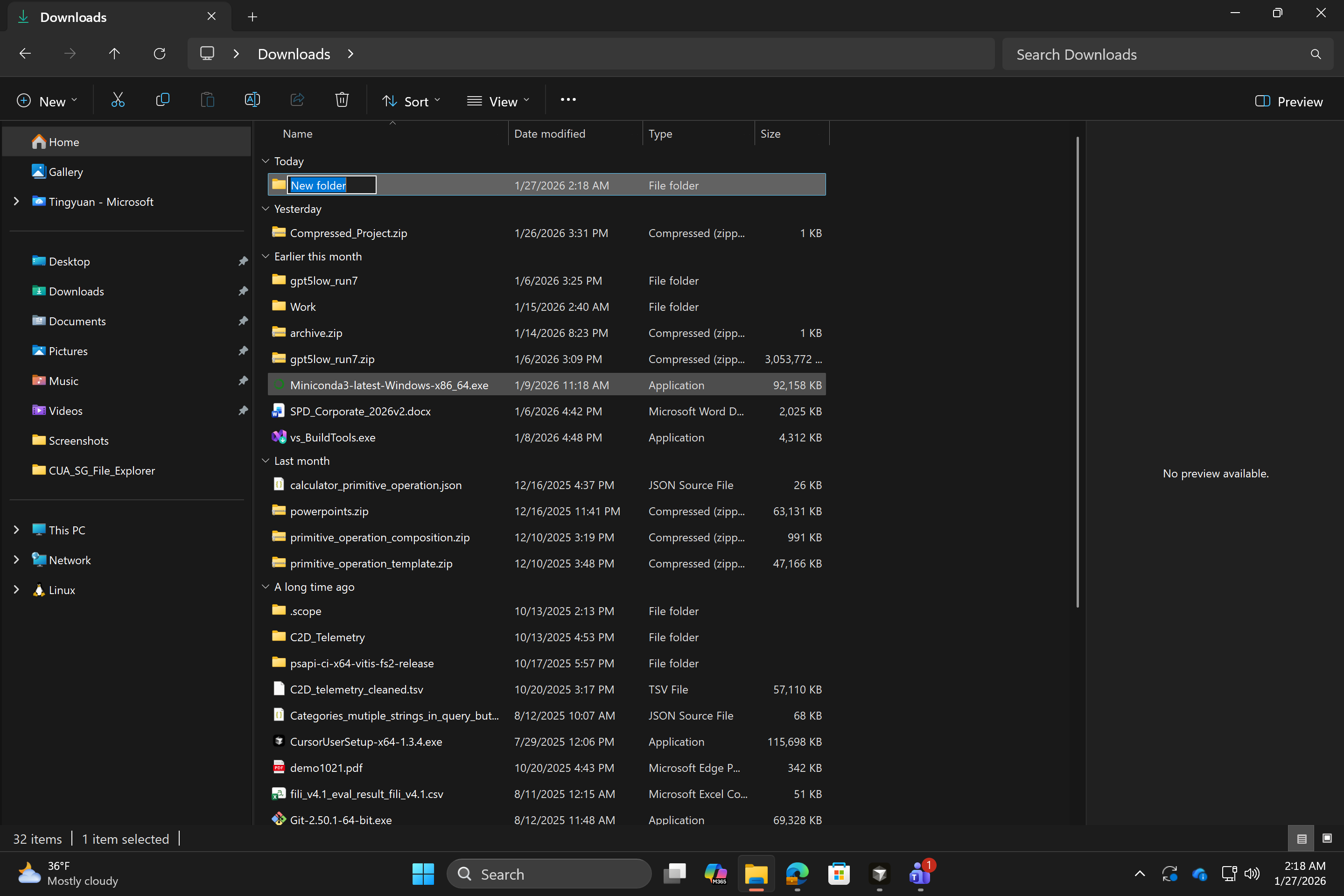}
\end{minipage}
\hfill
\begin{minipage}{0.58\linewidth}
\small
\textbf{Step 4: \texttt{HotKeyAction}}\\
\textbf{Reasoning.} Create new folder.\\
\textbf{Argument Instantiation.}
\begin{itemize}[leftmargin=1em]
    \item \texttt{keys:} [`ctrl', `shift', `n']
\end{itemize}
\end{minipage}

\vspace{0.4em}

\begin{minipage}{0.32\linewidth}
    \centering
    \includegraphics[width=\linewidth]{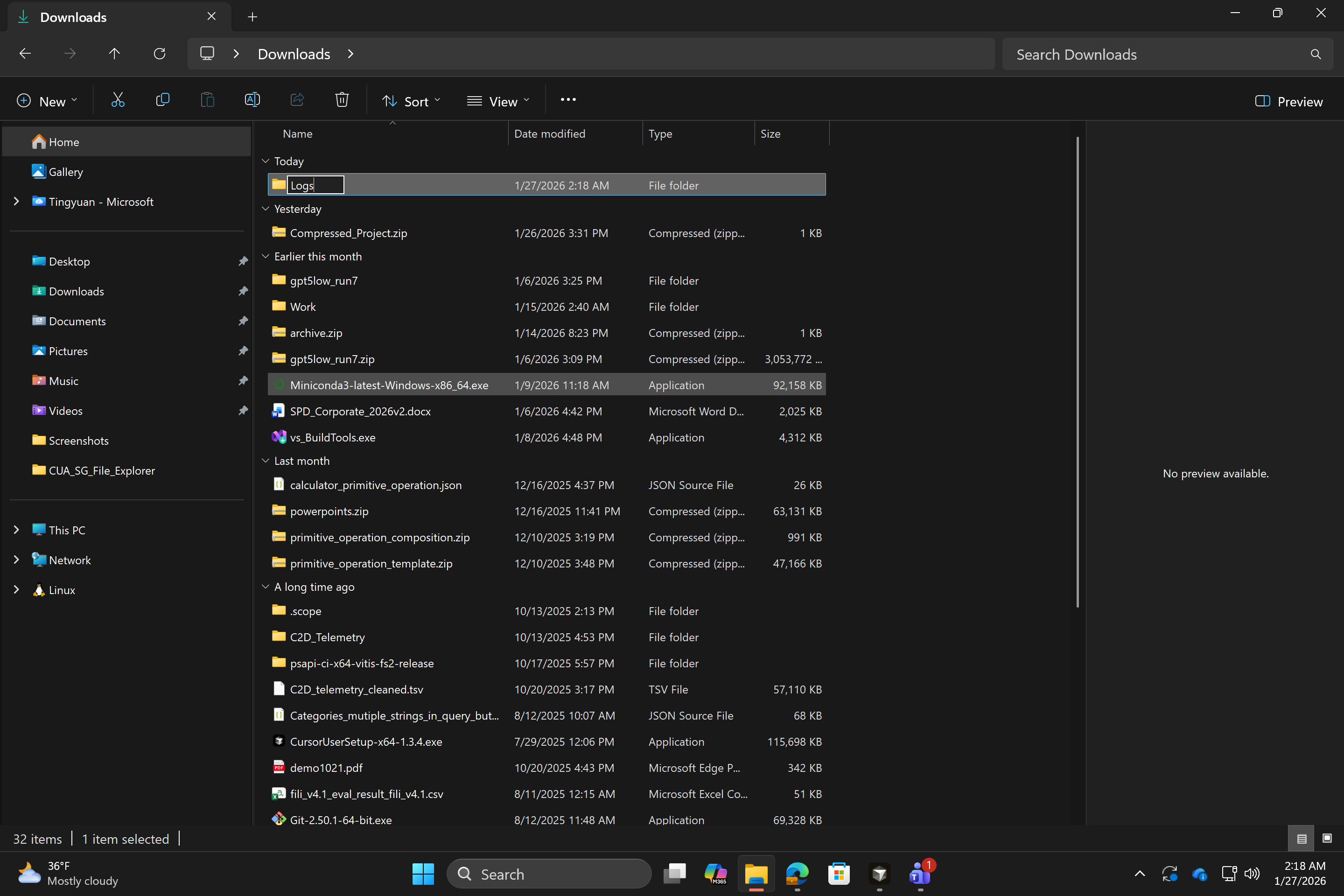}
\end{minipage}
\hfill
\begin{minipage}{0.58\linewidth}
\small
\textbf{Step 5: \texttt{TypeAction}}\\
\textbf{Reasoning.} Set folder name.\\
\textbf{Argument Instantiation.}
\begin{itemize}[leftmargin=1em]
    \item \texttt{input\_mode:} `copy\_paste'
    \item \texttt{text:} `Logs'
\end{itemize}
\end{minipage}

\vspace{0.4em}

\begin{minipage}{0.32\linewidth}
    \centering
    \includegraphics[width=\linewidth]{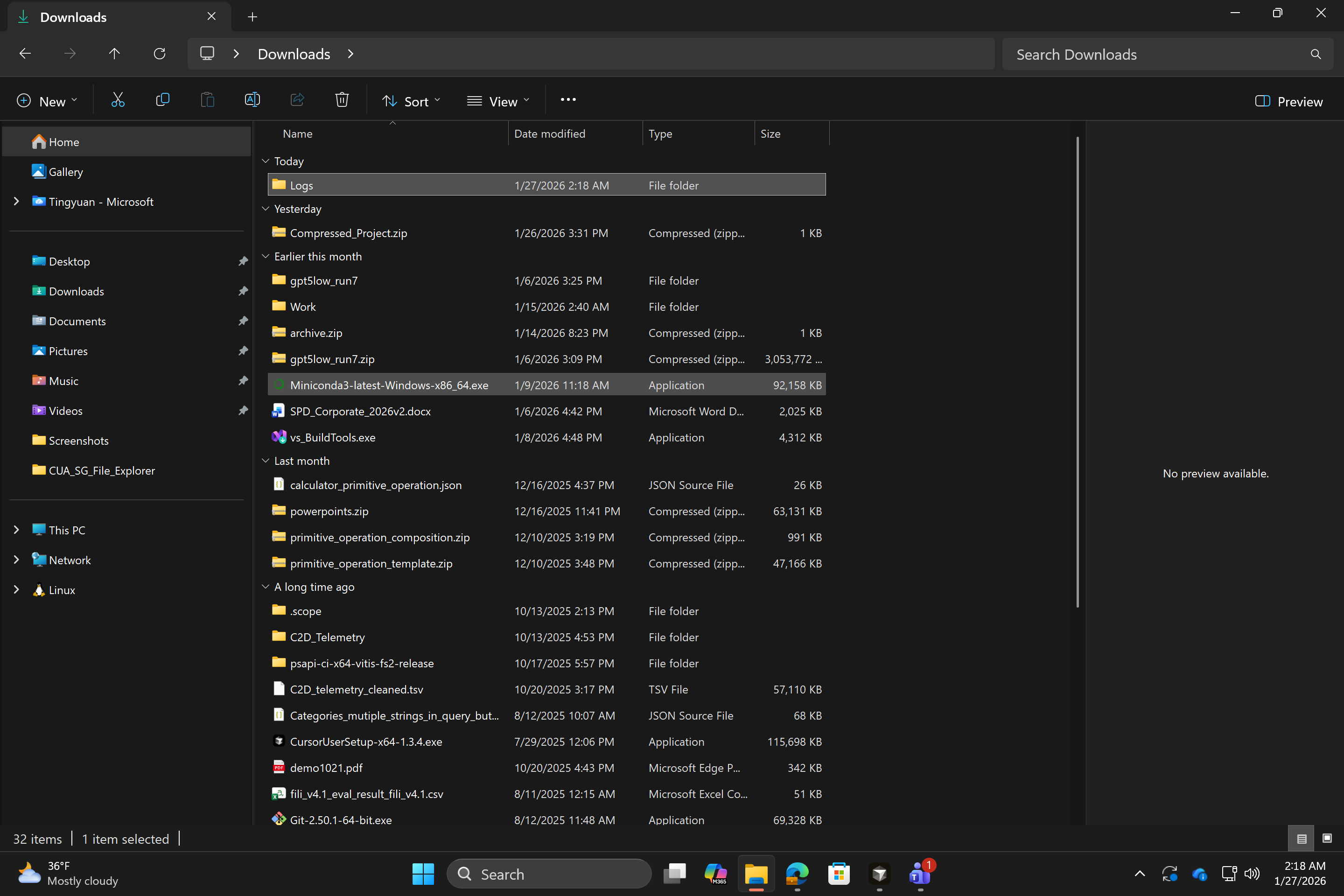}
\end{minipage}
\hfill
\begin{minipage}{0.58\linewidth}
\small
\textbf{Step 6: \texttt{HotKeyAction}}\\
\textbf{Reasoning.} Confirm new folder.\\
\textbf{Argument Instantiation.}
\begin{itemize}[leftmargin=1em]
    \item \texttt{key:} `Enter'
\end{itemize}
\end{minipage}
\hrule
\vspace{0.4em}

\small
\textbf{Outcome.} The agent successfully completed the skill.

\textbf{Key Insight.} The skill uses hotkey shortcuts in File Explorer to create and rename a folder, reducing failures from visual grounding, and only calling the grounding model to predict coordinates when needed.
\end{minipage}
\end{figure}

\begin{figure}[H]
\vspace{-0.5em}
\centering
\begin{minipage}{0.95\linewidth}
\paragraph{Case Study: Skill: \texttt{PowerPointSetTextFontColor}.}
\small

\textbf{Task:} Set the font color of selected text box to Light Blue.
\vspace{0.5em}
\hrule
\vspace{0.5em}

\begin{minipage}{0.4\linewidth}
    \centering
    \includegraphics[width=\linewidth]{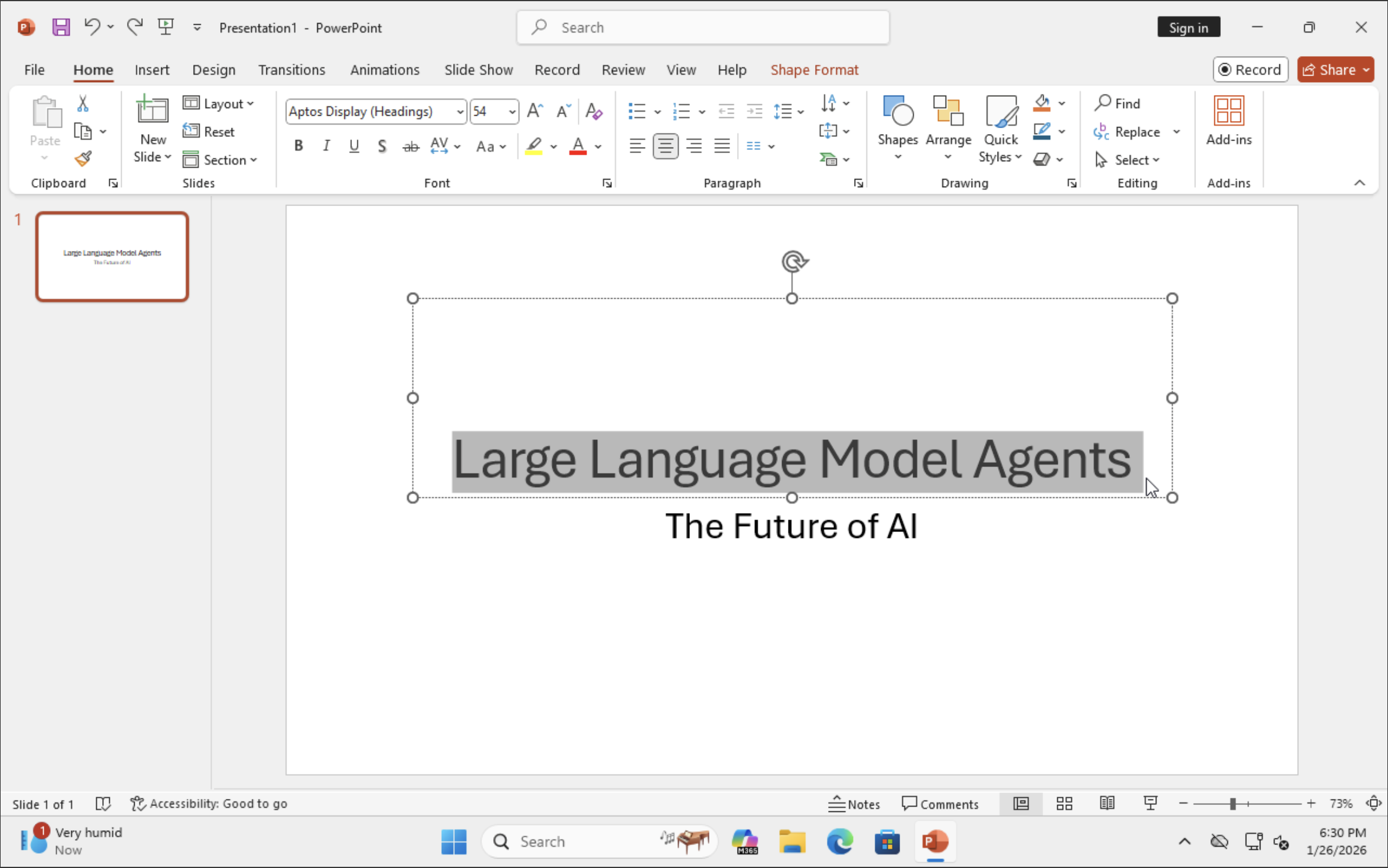}
\end{minipage}
\hfill
\begin{minipage}{0.5\linewidth}
\small
\textbf{Step 1: HotKeyAction}\\
\textbf{Reasoning.} Press Alt + H to switch to the Home tab.\\
\textbf{Argument Instantiation.}
\begin{itemize}[leftmargin=1em]
    \item \texttt{keys}: [`alt', `h'].
\end{itemize}
\end{minipage}

\begin{minipage}{0.4\linewidth}
    \centering
    \includegraphics[width=\linewidth]{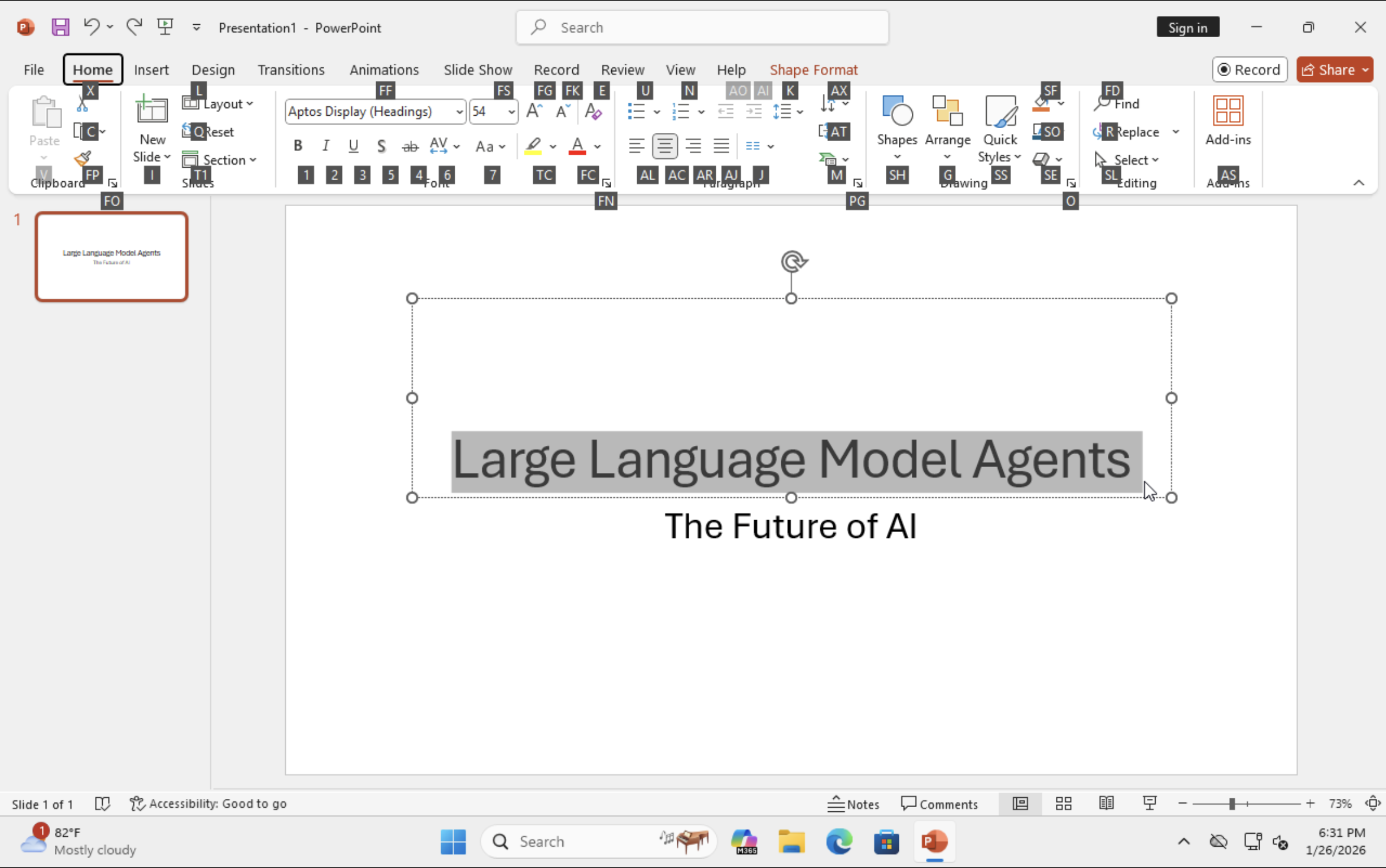}
\end{minipage}
\hfill
\begin{minipage}{0.5\linewidth}
\small
\textbf{Step 2: \texttt{PressKeyAction}}\\
\textbf{Reasoning.}
Press F and C to open the font color dropdown menu. First press F. \\
\textbf{Argument Instantiation.}
\begin{itemize}[leftmargin=1em]
    \item \texttt{key:} `f'
\end{itemize}
\end{minipage}

\begin{minipage}{0.4\linewidth}
    \centering
    \includegraphics[width=\linewidth]{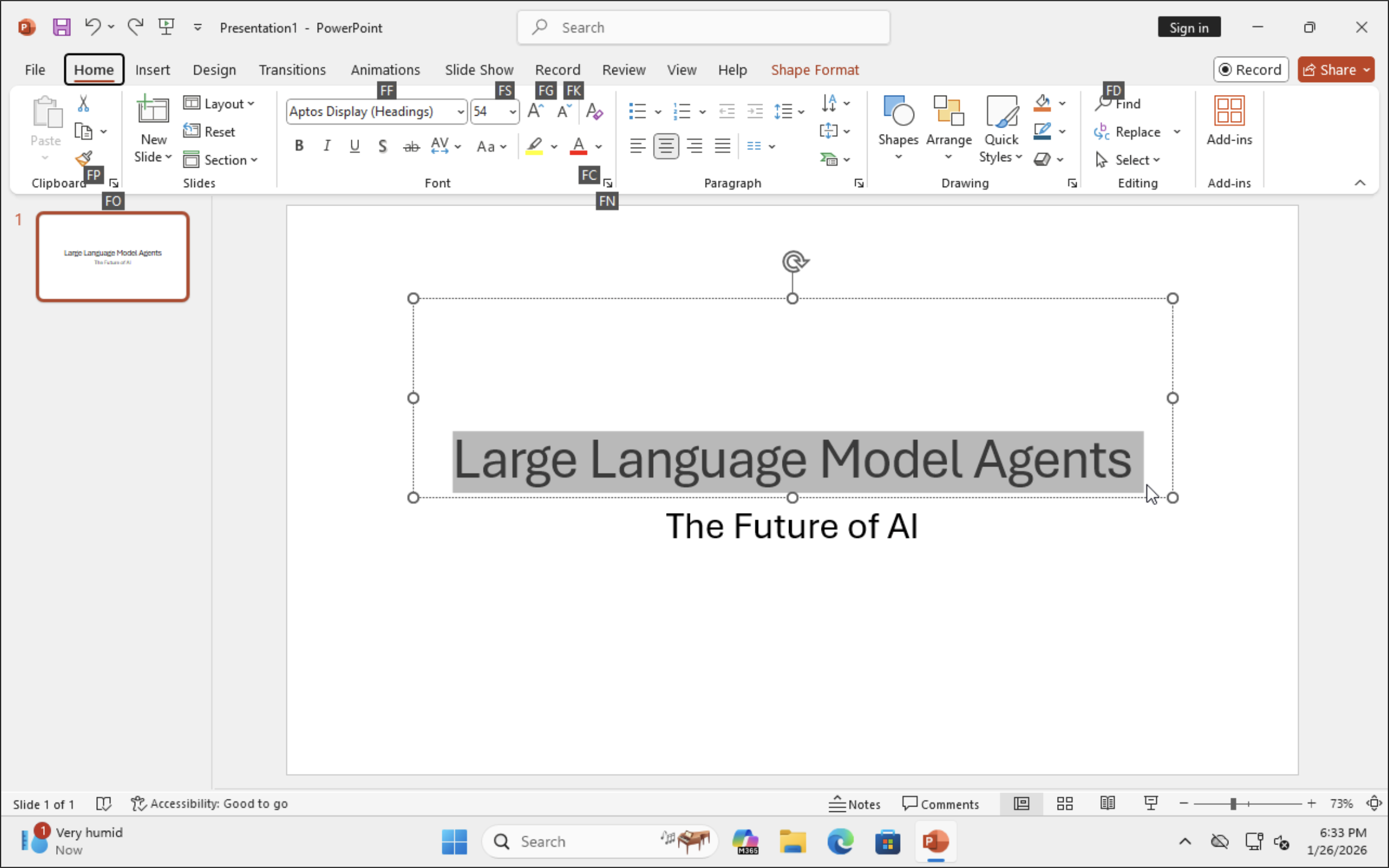}
\end{minipage}
\hfill
\begin{minipage}{0.5\linewidth}
\small
\textbf{Step 3: \texttt{PressKeyAction}}\\
\textbf{Reasoning.}
Then press C to complete the opening of the font color dropdown menu. \\
\textbf{Argument Instantiation.}
\begin{itemize}[leftmargin=1em]
    \item \texttt{key:} `c'
\end{itemize}
\end{minipage}

\begin{minipage}{0.4\linewidth}
    \centering
    \includegraphics[width=\linewidth]{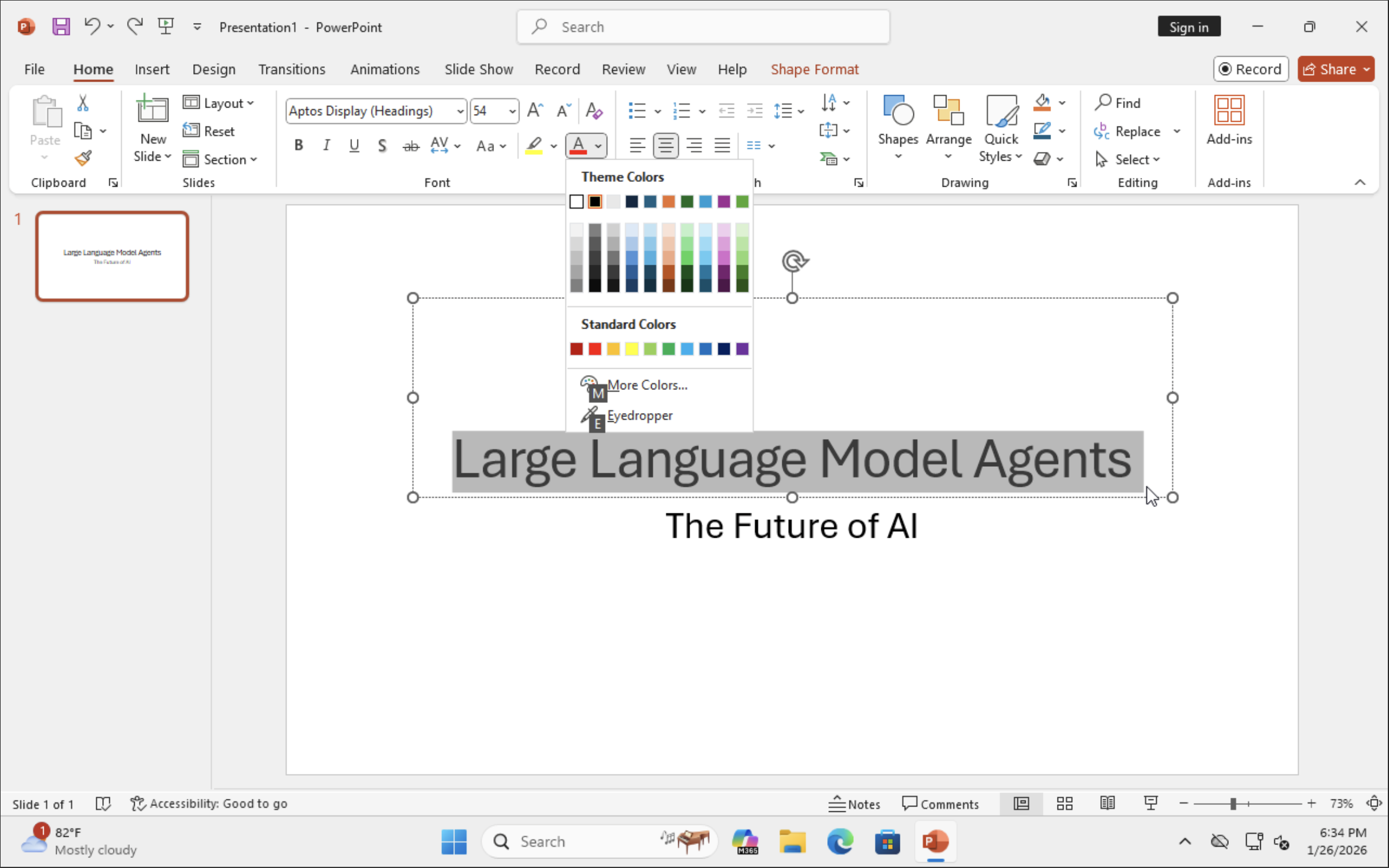}
\end{minipage}
\hfill
\begin{minipage}{0.5\linewidth}
\small
\textbf{Step 4: \texttt{SingleClickAction}}\\
\textbf{Reasoning.}
Click on the `Light Blue' color to set it for the selected text.\\
\textbf{Argument Instantiation.}
\begin{itemize}[leftmargin=1em]
    \item \texttt{Coordinate:} Call grounding model.
    \item \texttt{Button:} Left.
\end{itemize}
\end{minipage}

\begin{minipage}{0.4\linewidth}
    \centering
    \includegraphics[width=\linewidth]{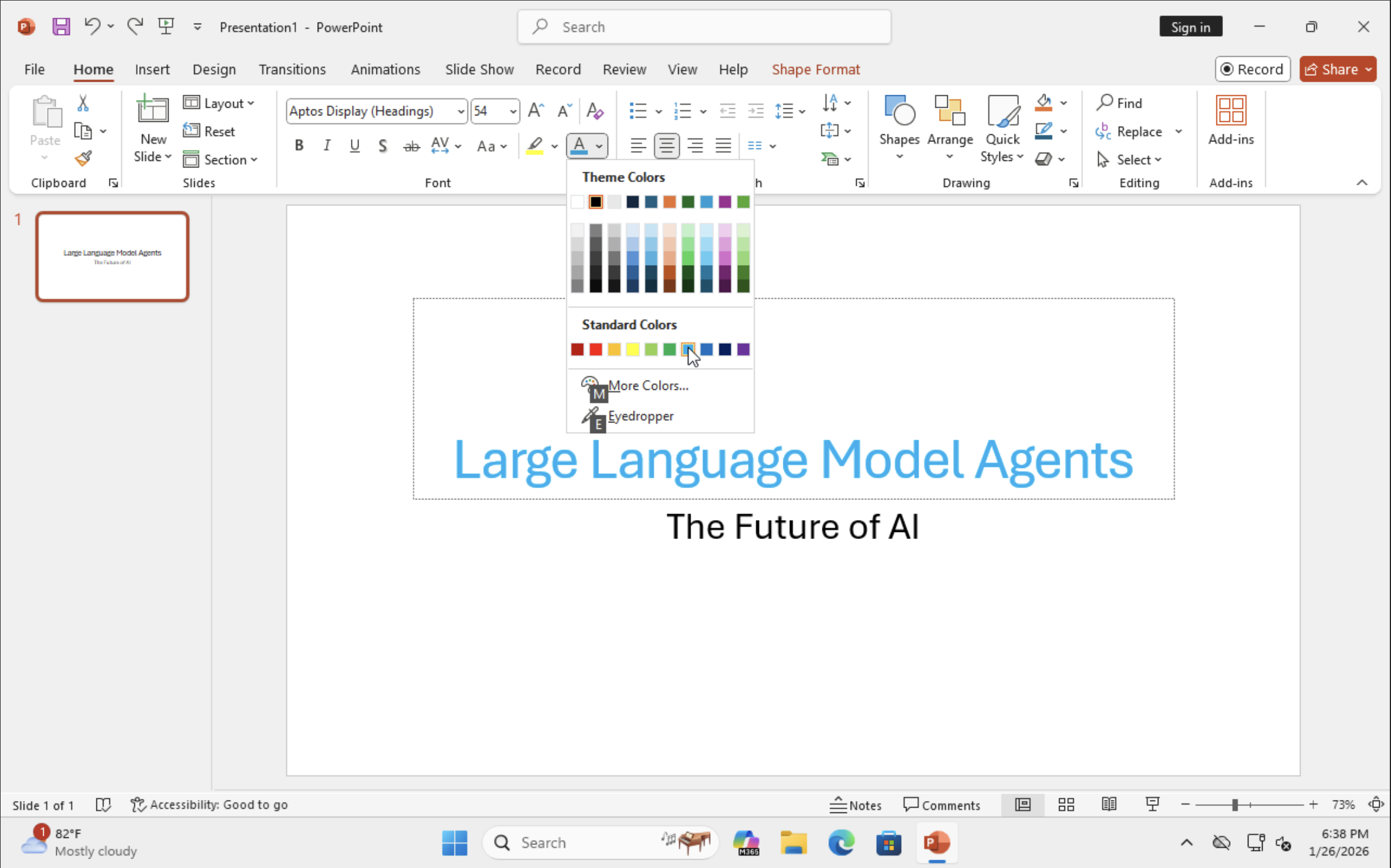}
\end{minipage}
\hfill
\begin{minipage}{0.5\linewidth}
\small
\textbf{Step 5: \texttt{HotKeyAction}}\\
\textbf{Reasoning.} Press Escape to close the font color dropdown menu. \\
\textbf{Argument Instantiation.}
\begin{itemize}[leftmargin=1em]
    \item \texttt{keys:} [`esc'].
\end{itemize}
\end{minipage}

\hrule
\vspace{0.4em}

\small
\textbf{Outcome.} The agent successfully completed the skill.

\textbf{Key Insight.} The skill uses hotkey shortcuts in PowerPoint to navigate in the ribbon, which reduces the possible failures from visual grounding, and only call grounding model to predict coordinates when needed.
\end{minipage}
\vspace{-1em}
\end{figure}

\newpage
\section{CUA Execution Graph}\label{appendix:cua_execution_graph}

\begin{figure}[H]
	\centering
	\includegraphics[height=0.85\textheight]{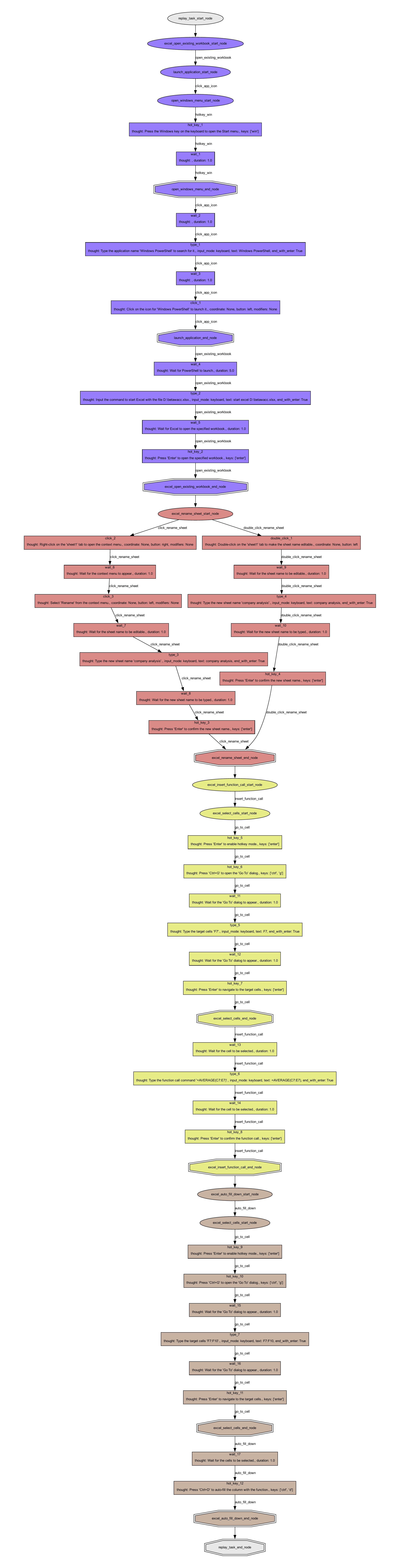}
	\caption{
		\textbf{CUA Task Execution Graph Example for Excel.}
		Open the file \texttt{betawacc.xlsx}, rename \texttt{Sheet1} to \texttt{company analysis}, and compute the \texttt{Average} column.
	}
	\label{fig:excel_execution_graph}
\end{figure}

\begin{figure}[H]
	\centering
	\includegraphics[height=0.85\textheight]{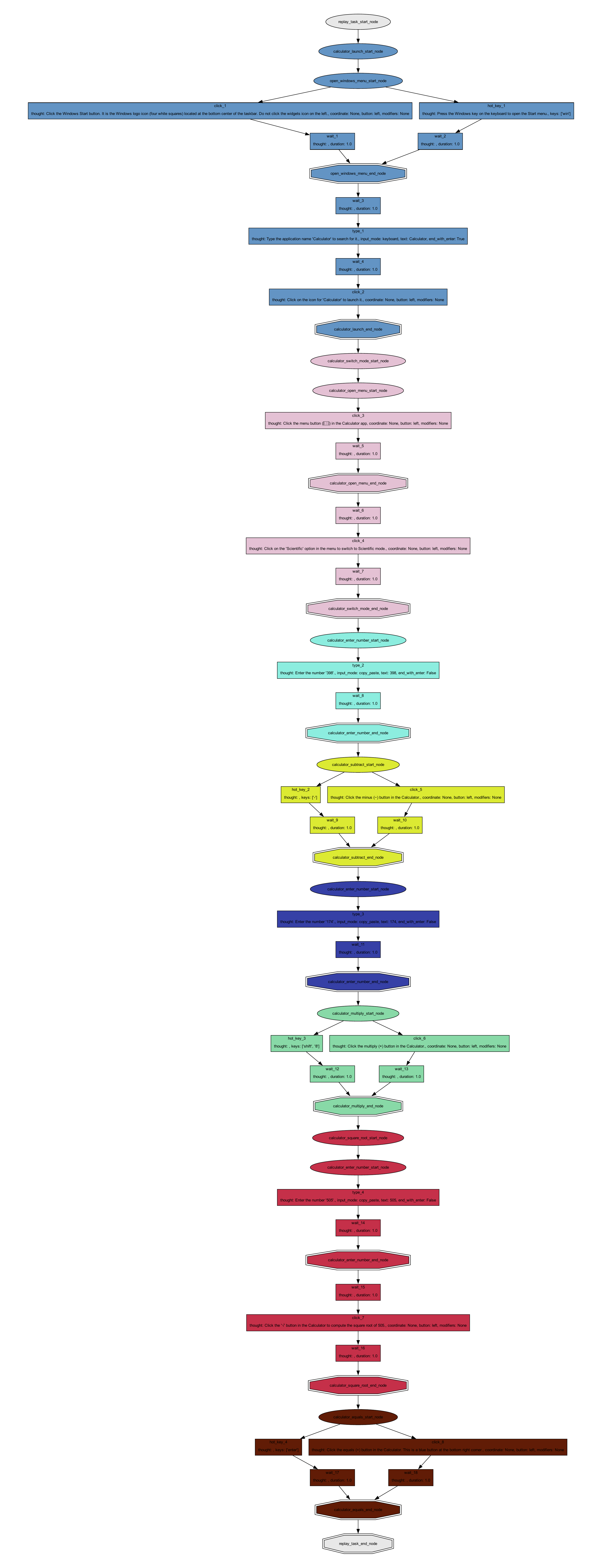}
	\caption{
		\textbf{CUA Task Execution Graph Example for Calculator.} Instruction: Calculate $398-174\times \sqrt{505}$.
	}
	\label{fig:excel_execution_graph}
\end{figure}

\end{document}